\documentclass{article}
\usepackage{iclr2027_conference,times}
\usepackage{amssymb} 
\usepackage{booktabs}
\usepackage{mathtools}
\usepackage{float}
\usepackage{hyperref}
\usepackage{cleveref}
\usepackage{titlesec}
\usepackage{tikz}
\usetikzlibrary{positioning,arrows.meta,fit,shadows}
\usepackage[most]{tcolorbox}
\definecolor{11Color}{HTML}{0F2C59}
\definecolor{RefColor}{HTML}{0066CC}
\hypersetup{
    colorlinks=true,
    linkcolor=RefColor,
    citecolor=11Color,
    urlcolor=RefColor,
}
\usepackage[table]{xcolor}

\usepackage{amsmath,amsfonts,bm}

\def\eqref#1{equation~\ref{#1}}

\def\1{\bm{1}}

\DeclareMathAlphabet{\mathsfit}{\encodingdefault}{\sfdefault}{m}{sl}
\SetMathAlphabet{\mathsfit}{bold}{\encodingdefault}{\sfdefault}{bx}{n}

\usepackage{adjustbox}
\usepackage{pifont} 
\usepackage{amsmath}
\usepackage{amssymb}
\usepackage{amsfonts}
\usepackage{mathtools}
\usepackage{bm}
\usepackage{bbm}

\usepackage{booktabs}
\usepackage{multirow}
\usepackage{multicol}
\usepackage{array}
\usepackage{makecell}
\usepackage{tabularx}

\usepackage{graphicx}
\usepackage{subcaption}
\usepackage{wrapfig}

\usepackage[linesnumbered,ruled,vlined]{algorithm2e}
\SetKwInput{Input}{Input}
\SetKwInput{Output}{Output}
\DontPrintSemicolon

\usepackage{enumitem}

\usepackage{xspace}
\usepackage{xcolor}
\usepackage{microtype}
\usepackage{hyperref}
\usepackage{url}

\definecolor{BlueGreen}{RGB}{0,130,120}
\definecolor{RedOrange}{RGB}{220,90,50}
\definecolor{PronounColor}{HTML}{238F8D}  
\definecolor{SuitcaseColor}{HTML}{D94F5C} 
\definecolor{TrophyColor}{HTML}{2478A8}   
\definecolor{E8F1FF}{HTML}{E8F1FF} 

\newcommand{\casebox}[3]{%
\begin{tcolorbox}[
  enhanced, breakable,
  colback=#1!4, colframe=#1!60!black,
  boxrule=0.6pt, arc=2.5pt,
  left=6pt, right=6pt, top=6pt, bottom=5pt,
  fonttitle=\small\bfseries\sffamily,
  coltitle=white, colbacktitle=#1!60!black,
  title={#2}
]
#3
\end{tcolorbox}}

\newcommand{\good}[1]{$_{\color{BlueGreen}\downarrow #1}$}
\newcommand{\bad}[1]{$_{\color{RedOrange}\uparrow #1}$}
\newcommand{\accgood}[1]{$_{\color{BlueGreen}\uparrow #1}$}
\newcommand{\accbad}[1]{$_{\color{RedOrange}\downarrow #1}$}
\newcommand{\modelname}[1]{\textbf{\textsc{#1}}}

\newcommand{\hlfirst}[1]{\colorbox[HTML]{CFE2FF}{#1}}  
\newcommand{\hlsecond}[1]{\colorbox[HTML]{E8F1FF}{#1}} 

\newcommand{\csA}[1]{\colorbox[HTML]{3465AF}{\textcolor{white}{#1}}}   
\newcommand{\csB}[1]{\colorbox[HTML]{3868B1}{\textcolor{white}{#1}}}   
\newcommand{\csC}[1]{\colorbox[HTML]{4371B5}{\textcolor{white}{#1}}}   
\newcommand{\csD}[1]{\colorbox[HTML]{5C84C0}{\textcolor{white}{#1}}}   
\newcommand{\csE}[1]{\colorbox[HTML]{5D85C1}{\textcolor{white}{#1}}}   
\newcommand{\csF}[1]{\colorbox[HTML]{6E92C8}{\textcolor{white}{#1}}}   
\newcommand{\csG}[1]{\colorbox[HTML]{7497CA}{\textcolor{white}{#1}}}   
\newcommand{\csH}[1]{\colorbox[HTML]{D9E5F5}{\textcolor{black}{#1}}}   
\newcommand{\csI}[1]{\colorbox[HTML]{DFEAF8}{\textcolor{black}{#1}}}   
\newcommand{\csJ}[1]{\colorbox[HTML]{E5EEFA}{\textcolor{black}{#1}}}   
\newcommand{\csK}[1]{\colorbox[HTML]{EAF3FD}{\textcolor{black}{#1}}}   
\newcommand{\csL}[1]{\colorbox[HTML]{F0F7FF}{\textcolor{black}{#1}}}   

\newcommand{\goodc}{\textcolor{green!60!black}{\ding{51}}}
\newcommand{\badc}{\textcolor{red!70!black}{\ding{55}}}
\newcommand{\partc}{\textcolor{orange!80!black}{$\sim$}}
\title{Write Back the $\Delta$: Revisiting the Same Tokens with Fresh  Representations}

\author{
Wencheng Ye$^{1,*}$,
Anning Hu$^{2,*}$,
Xiangdong Zhang$^{2}$,
Tianyi Wang$^{1}$,
Yikang Li$^{3}$,
Hengyu Jin$^{1}$,\\
\textbf{Bing Li}$^{1}$,
\textbf{Junchi Yan}$^{2,\dagger}$\\[1ex]
$^{1}$Tongji University,
$^{2}$Shanghai Jiao Tong University,
$^{3}$Shanghai Innovation Institute\\[0.5ex]
$^{*}$Equal contribution,
$^{\dagger}$Corresponding author
}

\begin{document}
\iclrfinalcopy
\pagestyle{plain}
\maketitle
\thispagestyle{plain}
\maketitle
\begin{abstract}
\vspace{-1em}
Transformers process information strictly forward through depth, preventing deeper computation from revisiting and refining earlier representations. To augment the standard forward pass, existing approaches either re-execute depth, incurring additional computation, or modify the residual stream using predefined directions, limiting their instance-level adaptation. Recently, inference-time feedback offers a direct mechanism for recycling endogenously produced computation by writing deeper residual states back to earlier layers, yet what should be fed back remains unclear. We argue that the depth increment $\Delta$, capturing newly accumulated computation between two layers, provides a more effective, composable, and scalable feedback signal than the full state. Building on this observation, we introduce \textbf{ReFlux}, a learnable feedback graph that dynamically selects and composes increment-carrying routes. ReFlux supports synchronous feedback to the same token and streaming feedback to subsequent tokens. Extensive experiments across various models, corpora, and benchmarks show that synchronous ReFlux consistently reduces perplexity across ten language-modeling corpora, and improves accuracy by \textbf{2.1--2.3 points}, with gains reaching \textbf{4.7 points} on multi-hop reasoning. Streaming ReFlux further retains most of these gains while preserving the base model's \textbf{1$\times$ theoretical backbone FLOPs}. These results establish ReFlux as an efficient paradigm for unlocking the latent computational potential of LLMs, allowing them to revisit the same tokens with fresh representations. Code implementation can be found at
\href{https://github.com/gooogleshanghai/reflux}{\texttt{https://github.com/gooogleshanghai/reflux}}.

\vspace{0.1em}

\end{abstract}
\noindent
\begin{minipage}[t]{0.30\linewidth}
    \section{Introduction}
\end{minipage}
\hfill
\begin{minipage}[t]{0.68\linewidth}
    \vspace{-2.2em}
    \begin{tcolorbox}[
        enhanced,
    frame hidden,
    borderline west={0.55pt}{0pt}{black},
    colback=white,
    left=9pt,
    right=7pt,
    top=2.5pt,
    bottom=2.5pt,
    boxsep=0pt,
    before skip=0pt,
    after skip=0pt
    ]
    \raggedleft
    {\fontsize{9.5pt}{10.2pt}\selectfont
    \itshape
    ``The real voyage of discovery consists not in seeking new landscapes,\\
    but in having new eyes.''\\[-1pt]
    \normalfont\small
    --- Marcel Proust, \textit{In Search of Lost Time}
    }
    \end{tcolorbox}
\end{minipage}

Some of our most valuable insights emerge only after deeper processing: a later distinction or connection can change how earlier evidence should be
interpreted. Intelligence, therefore depends not only on producing new information, but also on reusing it to reshape ongoing
interpretation, thereby \textbf{\textit{revising how we see with what we have discovered}}. Standard decoder-only Transformers lack a native path for this kind of reuse. Their residual stream accumulates
layer-wise updates in one direction through depth \citep{vaswani2017attention,he2016deep,elhage2021mathematical}. Later layers can build on earlier computation, but what they add cannot return to
condition earlier processing.

Existing approaches explore two ways to augment the standard forward computation.  Recurrent and looped Transformers \textit{re-execute depth}, obtaining refinement through
specialized training or additional inference computation \citep{dehghani2018universal,zhu2025scaling,bae2026mixture}. Activation
steering instead \textit{edits the residual stream}, but typically relies on directions extracted in advance from external data or predefined
behaviors \citep{turner2023steering,li2023inference,rimsky2024steering,o2024steering}, limiting instance-level adaptation
\citep{braun2025understanding,jiang2026global}. Rather than recomputing depth or applying a preset correction, we advocate for a complementary axis of reuse: \textbf{\textit{recycling the computation the model has already produced endogenously.}}


Recirculation moves closer to this goal by writing a deeper residual state back to an earlier layer in a frozen language model
\citep{mozer2026recirculation}. However, because the residual stream is cumulative, the deep state contains both information already
present at the earlier layer together with newly accumulated computation, making feedback increasingly redundant as routes accumulate. This raises a sharper question: \textbf{\textit{should feedback replay where the model arrived, or only what depth added?}}


\begin{figure*}[t]
    \centering
    \includegraphics[width=0.99\textwidth]{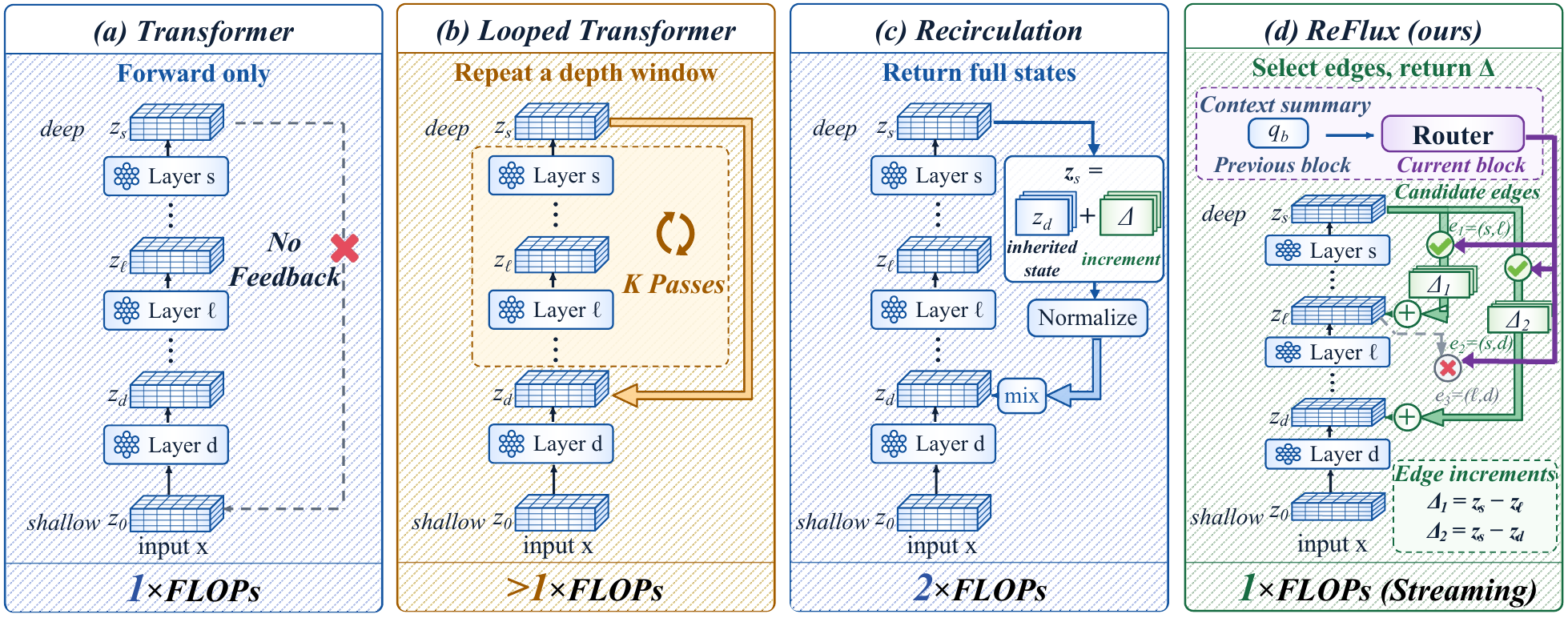}
    \caption{Revisiting earlier representations with new computation. Looped Transformers re-execute depth, while Recirculation returns full states to earlier layers.
    ReFlux instead isolates the newly accumulated computation and selectively routes depth increments $\Delta$ through a learned
    feedback graph, enabling streaming feedback without increasing the backbone's theoretical FLOPs.}
    \label{fig:motivation}
\end{figure*}
We answer by using newly added computation as a cleaner unit of feedback. For a deeper state $z_s$ and an earlier state $z_d$, we define the \textbf{depth increment} as
\(\boldsymbol{\Delta}_{d\rightarrow s}=z_s-z_d\),
which captures the residual computation accumulated between the two layers. Controlled studies show that full-state feedback
saturates as more feedback is introduced, whereas increment feedback continues to benefit from additional routes and rounds.
Moreover, increments from different layer pairs provide complementary gains. Feedback thus becomes an allocation problem: \textit{\textbf{which}}
increments should be reused, \textit{\textbf{where}} should they be written, and with \textit{\textbf{what}} strength for the current context?



To make this form of reuse a first-class operation, we propose \textbf{ReFlux}, a learnable sparse feedback graph over frozen Transformers (Fig.~\ref{fig:motivation}). Each edge
carries an increment from a deeper source to a shallower target, while a lightweight router selects the
active edges and their strengths conditioned on the current context. The same graph supports two execution schedules: \textbf{synchronous} execution revisits a token using its own
deeper computation, whereas \textbf{streaming} execution carries the interpretation accumulated over the processed prefix forward to shape
subsequent tokens within a single traversal.




Across five models from two families, ten language-modeling corpora, and eight reasoning benchmarks, synchronous ReFlux consistently improves language modeling performance, yielding systematic perplexity reductions over the base models and prior feedback baselines.
 It also improves average reasoning accuracy by \textbf{2.1--2.3 points}. Gains are largest on multi-hop
tasks, averaging \textbf{4.7 points} where answers require evidence to be composed across distributed information. Streaming ReFlux retains most of these
improvements at the base model's\textbf{ 1$\times$ theoretical backbone FLOPs }and only \textbf{1.02--1.08$\times$ decoding runtime}. Routing
analyses further show that ReFlux remains sparse while adapting its selected support to input difficulty and task family.

Together, ReFlux turns depth from a one-way computation path into an input-adaptive source of reusable updates, allowing the model to take a second look at the same tokens with information revealed later. Our contributions are
summarized as follows:

\begin{itemize}[leftmargin=2em,itemsep=-0.1em]
\item[\ding{182}] \textbf{\textit{Discover}.} We identify the depth increment $\Delta$ as an effective and scalable feedback signal, turning newly accumulated computation into a reusable and composable unit that scales across different dimensions.

\item[\ding{183}] \textbf{\textit{Build.}} We propose ReFlux, a learnable feedback graph over frozen Transformers that dynamically composes increment-carrying routes, with a streaming execution scheme that preserves the base model's $1\times$ theoretical backbone FLOPs.

\item[\ding{184}] \textbf{\textit{Validate.}} We demonstrate consistent gains across five models, ten corpora, and eight benchmarks, including improved language modeling and a 4.7 point gain in multi-hop reasoning, while streaming execution preserves most of these benefits at only $1.02$–$1.08\times$ decoding runtime.
\end{itemize}

\section{Related Work}
\subsection{Recurrent and Looped Transformers}

Adding effective depth through layer iteration is a long-standing direction in language-model research. \citet{dehghani2018universal} introduced the Universal Transformer, which shares a block across recurrent steps, while \citet{geiping2026scaling} extend this paradigm to billion-parameter models. Looped transformers \citep{yang2024looped,giannou2023looped} and looped language models \citep{zhu2025scaling,kohli2026loop} use recurrent depth for algorithmic computation and parameter efficiency, while SMELT \citep{wang2026smelt} scales compute-matched looping to 54B-parameter MoE models. The same principle extends to latent reasoning, with Coconut \citep{hao2024training}, PonderLM \citep{zeng2025ponderlm}, implicit chain-of-thought \citep{deng2023implicit}, Quiet-STaR \citep{zelikman2024quiet}, and TRM \citep{jolicoeur2025less} exploring recurrent or recursive latent computation. Adaptive computation further learns where or when to allocate computation, as explored by ACT \citep{graves2016adaptive}, PonderNet \citep{banino2021pondernet}, MoD \citep{raposo2024mixture}, and MoR \citep{bae2026mixture}. These approaches either train pretrained models to introduce recurrence or feedback, as in Encode--Think--Decode \citep{koishekenov2025encode} and FBT \citep{wang2026full}, or operate training-free by re-executing existing layers, as in TF-Loop \citep{chen2026training} and CoLa \citep{li2025skip}. Thus, adding depth to a checkpoint still requires either training or extra inference FLOPs. ReFlux  instead treats increments already computed by the network as reusable payloads and turns their selection into a routing problem, enabling effective depth without additional backbone FLOPs in streaming.

\subsection{Inference-Time Interventions on Residual Stream}
In parallel, another family of methods edits the residual stream directly at inference time. Activation steering adds behavior-specific directions to internal activations, with prior work exploring directions derived from manually constructed signals \citep{turner2023steering}, labeled or paired behaviors \citep{li2023inference,ye2026riser,rimsky2024steering}, and representation features \citep{o2024steering}.  However, because these directions are typically extracted in advance, they may adapt poorly to individual inputs \citep{braun2025understanding}, suffer from limited generalization beyond the extraction distribution \citep{jiang2026global}, or induce undesirable shifts in activation distributions
\citep{stickland2024steering}. Patchscopes \citep{ghandeharioun2024patchscopes} re-enters the model with its own representation for readout, rather than feeding it back into the computation that produced it. More recently, Recirculation \citep{mozer2026recirculation} introduces a feedback mechanism that writes a deeper residual state back into a shallower stream. ReFlux changes what and how information is propagated: it feeds back the depth increment produced by the model itself through a learned feedback mechanism.

Extended related works and a comparison of ReFlux with these methods (Table~\ref{tab:positioning}) are  in Appendix~\ref{app:more_related}.

\section{THE $\Delta$: A Unit of New Computation}

\subsection{Deep Computation as Feedback}
Consider a decoder-only Transformer that repeatedly updates a shared residual stream. Let
$z_\ell(t) \in \mathbb{R}^{D}$ denote the residual stream for token $t$ after
layer $\ell$. Each block adds a layer-specific update:
\begin{equation}
z_{\ell+1}(t) = z_\ell(t) + F_\ell(z_\ell,t).
\label{eq:residual_update}
\end{equation}
Over depth, these updates can integrate broader context and form distinctions unavailable near the bottom of the stack. However, computation is
directional across depth: a shallow state can be refined by later blocks, but
later interpretations cannot directly revise earlier states. Once carried
forward, a block's computation only affects downstream computation, with no
route back to the earlier stream.

Recirculation introduces precisely such a feedback path in a frozen language
model. Given a source layer $s$ and a target layer $d$ with $s>d$, it writes
the deep residual state back to the target stream:\begin{equation}
\widetilde z_d(t)
= \beta z_d(t) + \alpha\,\mathcal{N}_{z_d(t)}\bigl(z_s(t)\bigr),
\qquad
\mathcal{N}_{h}(z)=
\frac{\lVert h\rVert_2}{\lVert z\rVert_2+\varepsilon}z.
\label{eq:recirculation}
\end{equation}
Here, $\alpha$ controls the feedback strength, $\beta$ retains the target
state, and $\mathcal{N}_{h}$ matches the source state to the target scale.
In Recirculation, Eq.~\ref{eq:recirculation} uses the full source state $z_s$, which raises a natural question: \emph{is the full state
the right signal to write back?}

\subsection{Uncovering the $\Delta$}
\begin{figure*}[t]
\centering
\begin{minipage}[c]{0.60\textwidth}
    \centering
     \hspace*{-0.035\textwidth}
    \includegraphics[width=\textwidth]{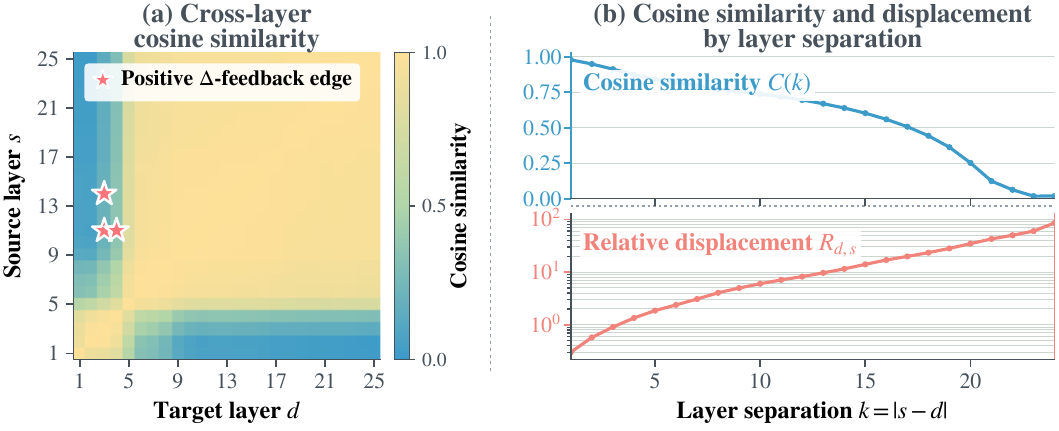}
\end{minipage}
\hfill
\begin{minipage}[c]{0.39\textwidth}
    \caption{Cross-depth geometry of Gemma3-1B residual states.
    (a) Pairwise cosine similarity between $z_d$ and $z_s$;
    stars mark top-three positive source-target pairs. (b) Cosine similarity and
    norm-relative displacement versus layer separation $k$. Adjacent
    states are similar, while successive layerwise rewrites accumulate
    into a larger change over longer depth intervals.}
    \label{fig:cross_depth_geometry}
\end{minipage}
\end{figure*}

To answer this question, we first examine what is actually new at each depth. Specifically, we characterize how residual states change across depth under the same sampling protocol as \citet{mozer2026recirculation}, using 1,000
fixed-length windows of 1,024 tokens from the C4 validation split
\citep{raffel2020exploring}. For each window, we
record the block residual stream and compute the pairwise cosine similarity
between depths $d$ and $s$:
\[
C_{d,s}
=
\mathbb{E}_{t}
\left[
\cos\bigl(z_d(t),z_s(t)\bigr)
\right].
\]
We average over token positions within each subset and then aggregate across
subsets. We also measure the norm-relative displacement
$R_{d,s}\coloneqq\mathbb{E}_{t}[\lVert z_s(t)-z_d(t)\rVert_2/
(\lVert z_d(t)\rVert_2+\varepsilon)]$. For layer separation $k$, we use
the depth-averaged profiles
\[
C(k)\coloneqq\frac{1}{L-k}\sum_{\ell=1}^{L-k}C_{\ell,\ell+k},
\qquad R(k)\coloneqq\frac{1}{L-k}\sum_{\ell=1}^{L-k}R_{\ell,\ell+k}.
\]
Here $L$ is the number of analyzed block states. Fig.~\ref{fig:cross_depth_geometry}(b) shows that adjacent
states are highly similar, yet the layerwise rewrites
accumulate over depth.
Fig.~\ref{fig:cross_depth_geometry}(a) gives the corresponding layer-pair view. The three marked best
pairs were obtained from an independent  sweep
over source-target choices. They lie away
from the high-similarity diagonal and connect shallow targets to substantially
deeper sources, indicating that useful feedback emerges after multiple
layerwise rewrites have accumulated.

The deeper state is therefore not just a later copy of the target state, it
mixes inherited content with the computation accumulated between the two
depths. We therefore distill
the depth-wise increment
\begin{equation}
\Delta_{d\rightarrow s}(t)
\coloneqq z_s(t)-z_d(t).
\label{eq:delta_increment}
\end{equation}
To isolate its role, we compare three matched payloads: the full source state
$z_s$, the target state $z_d$, and the increment in
Eq.~\ref{eq:delta_increment}. Within each model, the frozen model,
source--target pair, injection positions, feedback schedule, feedback strength,
and scale protocol are held fixed, only the payload changes. We report
perplexity, relative perplexity change, and the mean token-level NLL
reduction. As Table~\ref{tab:payload_decomposition} shows, the
increment is consistently best on architectures with different sizes and normalization
schemes. Writing back the target state provides little to no improvement, while the full source state yields smaller gains than the isolated increment, indicating that the feedback benefit lies not in the state itself, but in the change induced by deeper computation.

\begin{table}[t]
\centering
\caption{Matched-payload decomposition under architecture-specific stable
feedback protocols on C4, with 64 windows of 512 tokens for each model.
We use the best-performing source--target configuration. For Gemma3-1B, we use
$s{=}11\to d{=}4$, $\alpha{=}0.25$, $\beta{=}0.9$;
for Gemma3-4B, $s{=}18\to d{=}9$, $\alpha{=}0.15$, $\beta{=}0.9$; for Gemma3-12B, $s{=}35\to d{=}16$, $\alpha{=}0.15$, $\beta{=}1.0$; 
for Qwen3-4B, $s{=}26\to d{=}13$,
$\alpha{=}0.2$, $\beta{=}1.0$.  }
\label{tab:payload_decomposition}

\footnotesize
\setlength{\tabcolsep}{4pt}
\renewcommand{\arraystretch}{1.05}

\resizebox{\textwidth}{!}{%
\begin{tabular}{@{}lcc@{\qquad}cc@{\qquad}cc@{\qquad}cc@{}}
\toprule

&
\multicolumn{2}{c}{\modelname{Gemma3-1B}} &
\multicolumn{2}{c}{\modelname{Gemma3-4B}} &
\multicolumn{2}{c}{\modelname{Gemma3-12B}} &
\multicolumn{2}{c}{\modelname{Qwen3-4B}} \\

\cmidrule(lr){2-3}
\cmidrule(lr){4-5}
\cmidrule(lr){6-7}
\cmidrule(lr){8-9}

Payload
& PPL $\downarrow$ & $\Delta\log$PPL ($\times 10^{-2}$)  $\uparrow$
& PPL $\downarrow$ & $\Delta\log$PPL ($\times 10^{-2}$) $\uparrow$
& PPL $\downarrow$ & $\Delta\log$PPL ($\times 10^{-2}$) $\uparrow$
& PPL $\downarrow$ & $\Delta\log$PPL ($\times 10^{-2}$) $\uparrow$
\\

\midrule

None
& 30.772\good{0.00} & 0.00
& 23.693\good{0.00} & 0.00
& 42.973\good{0.00} & 0.00
& 25.562\good{0.00} & 0.00
\\

Full $z_s$
& 29.699\good{3.49} & +3.55
& 22.790\good{3.81} & +3.89
& 41.462\good{3.52} & +3.58
& 24.626\good{3.66} & +3.73
\\

Base $z_d$
& 31.110\bad{1.10} & -1.09
& 23.899\bad{0.87} & -0.87
& 43.451\bad{1.11} & -1.11
& 25.599\bad{0.14} & -0.14
\\

$\Delta_{d\to s}$
& \hlfirst{\textbf{29.210\good{\textbf{5.08}}}}
& \hlfirst{\textbf{+5.21}}
&
\hlfirst{\textbf{21.856\good{\textbf{7.75}}}}
& \hlfirst{\textbf{+8.07}}
&
\hlfirst{\textbf{40.231\good{\textbf{6.38}}}}
& \hlfirst{\textbf{+6.59}}
&
\hlfirst{\textbf{24.030\good{\textbf{5.99}}}}
& \hlfirst{\textbf{+6.18}}

\\

\bottomrule
\end{tabular}}
\end{table}

\subsection{Scaling Beyond Saturation}
\label{sec:scaling}
We next ask whether this $\Delta$ advantage survives when feedback is scaled
along three axes: the number of rounds $r$, the graph width $k$, and the
context length $T$. Here, $r$ and $k$ denotes the number of feedback times and active feedback edges. The round and width sweeps evaluate MMLU accuracy
\citep{hendrycks2020measuring} and BigPatent perplexity 
\citep{sharma2019bigpatent}, using 2,000 randomly sampled examples; the length
sweep measures mean $\Delta$PPL on 50 PG19 validation set books \citep{rae2019compressive}.
We first rank all source--target layer
pairs by validation improvement and use nested top-$k$ sets
$\mathcal{E}_k=\{e_1,\ldots,e_k\}$ for the width sweep. We evaluate
$r\in\{0,1,2,3,4,6\}$ feedback rounds and $k\in\{0,1,2,3,4,6\}$ active edges.
For the length sweep, we use
$T\in\{512,1{,}024,2{,}048,4{,}096,8{,}192\}$. Across the width and length
experiments, we keep the total feedback strength
$A_{\mathrm{total}}=\sum_{e\in\mathcal{E}}\alpha_e$ fixed at $0.5$. For the
width sweep, this strength is split equally across edges. We report the length result as
${\footnotesize
G_{\mathrm{PPL}}(T)
=100\left(1-\frac{P_{\mathrm{fb}}(T)}{P_{0}(T)}\right)}
$,
so values near zero indicate that feedback no longer changes perplexity
materially. 

\begin{figure*}[t]
\centering
\includegraphics[width=0.97\textwidth]{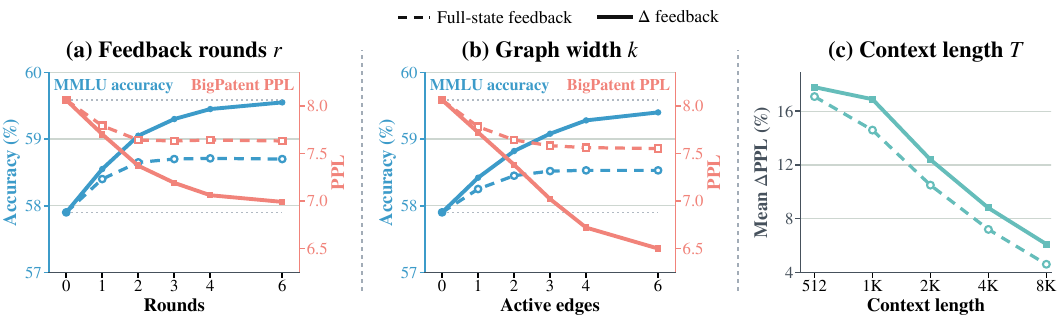}
\caption{Scaling behavior of feedback across three axes on Gemma3-4B.}
\label{fig:scaling}
\end{figure*}
\textbf{The round sweep} in Fig.~\ref{fig:scaling}(a) probes the \emph{depth}
of feedback, asking how much iterative refinement can be usefully applied.
Full-state feedback gives an early gain but levels off quickly, whereas
$\Delta$ feedback continues to improve over a wider range of $r$.  Repeated refinement is therefore more effective when each write carries newly
accumulated computation.

\textbf{The width sweep} in Fig.~\ref{fig:scaling}(b) probes the \emph{breadth}
of feedback, asking how many source--target routes can be combined. Here $k$
is the number of active edges, carrying the source state or depth increment. Full-state feedback saturates after a few edges, whereas
$\Delta$ continues to benefit from a wider graph, suggesting that increment payloads compose more cleanly across routes.

\textbf{The context sweep} in Fig.~\ref{fig:scaling}(c) probes how feedback
behaves as \emph{length} grows. As longer prefixes provide the backbone with
more usable context, the marginal headroom for a fixed-strength intervention
decreases, and the relative gain declines for both protocols. $\Delta$ retains more of its short-context benefit, with its performance decaying more slowly as context length increases.

Taken together, $\Delta$ feedback retains useful headroom when scaled in depth, breadth, and length. These axes imply different ways to expand feedback: increasing $r$ enables additional rounds of refinement but requires further stack traversals, whereas increasing $k$ introduces more increment-carrying routes without altering the refinement schedule. We therefore  focus on the compositional capacity of feedback routes at a fixed refinement depth, where increasing $k$ provides a natural and inexpensive benefit. A static intervention, however, fixes its routes and strengths in advance, leaving it unable to select and combine increments according to the current context. This motivates a more flexible and adaptive formulation in which the available feedback budget can be dynamically allocated across increment-carrying routes.

\section{ReFlux}
\label{sec:method}

\begin{figure}[t]
\centering
\includegraphics[width=\linewidth]{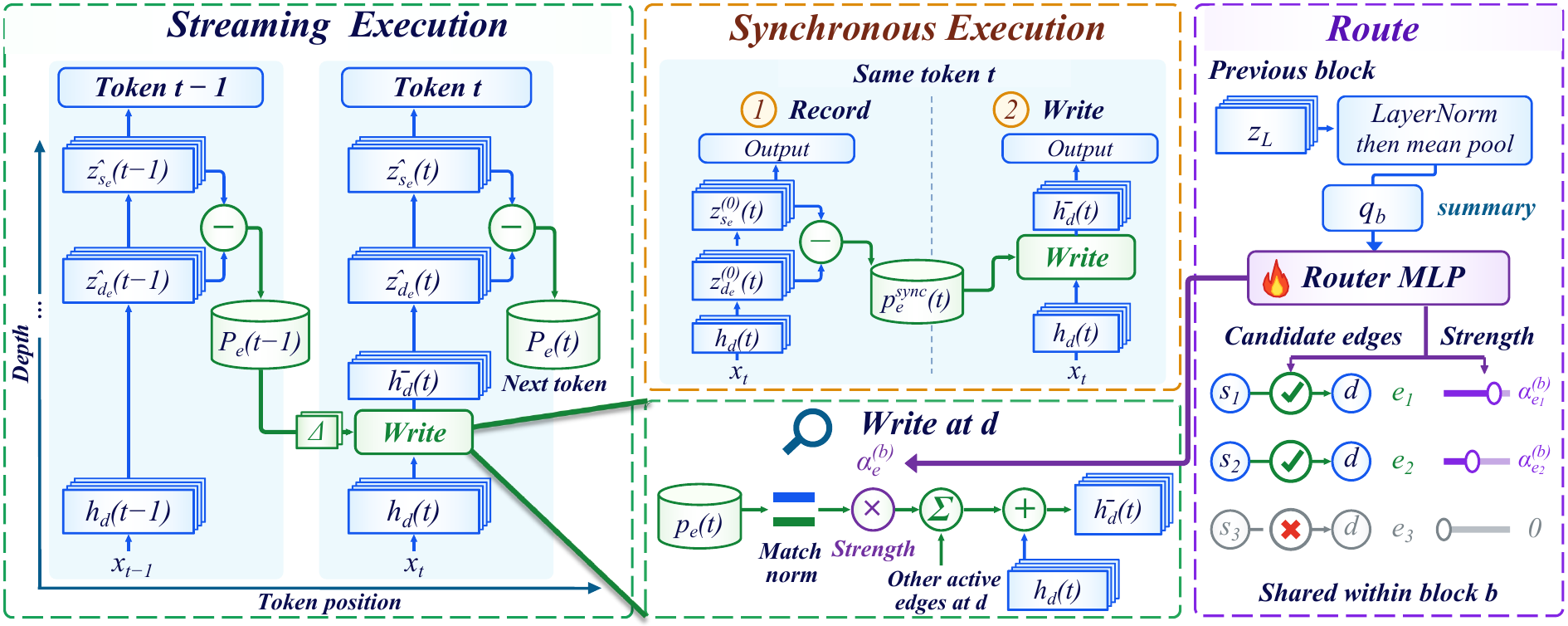}
\caption{Overview of ReFlux Architecture and Execution Schedules.}
\label{fig:method}
\end{figure}

ReFlux instantiates this idea as a feedback graph over a frozen decoder-only Transformer, as illustrated in Fig.~\ref{fig:method}. Candidate edges connect deeper source states to shallower target
states and carry the increments accumulated between the two depths. Let
$\mathcal{T}_b$ denote the token positions in routing block $b$. The router
produces one route for each block, shared by all $t\in\mathcal{T}_b$.

\subsection{Increment Edges}

Let the frozen Transformer have depth nodes $V=\{0,\ldots,L\}$, one node per
residual-stream depth. We define a candidate feedback graph
$\mathcal{G}_C=(V,\mathcal{C})$, where each directed edge
$e=(s_e,d_e)\in\mathcal{C}$ satisfies $s_e>d_e$ and connects a deeper source
depth to a shallower target depth. For a token position $t$, let $h_d(t)$
denote the residual state at target depth $d$ immediately before the write.
Given reference states $z^{\mathrm{ref}}$ recorded from a traversal, the
payload carried by edge $e$ is the depth increment
\begin{equation}
p_e(t)=\Delta_e(t)=z_{s_e}^{\mathrm{ref}}(t)-z_{d_e}^{\mathrm{ref}}(t).
\label{eq:edge_payload}
\end{equation}
The execution schedule determines which traversal supplies these reference
states. The edge therefore exposes the computation accumulated over the
interval $(d_e,s_e]$.

For block $b$, the router activates a subset
$\mathcal{E}^{(b)}\subseteq\mathcal{C}$. For target depth $d$, let
$\mathcal{I}_d^{(b)}=\{e\in\mathcal{E}^{(b)}:d_e=d\}$ denote its active incoming
edges. Given nonnegative edge strengths $\alpha_e^{(b)}$, ReFlux writes the
selected payloads additively into the target stream at position $t$:
\begin{equation}
\bar h_d(t)
=
h_d(t)
+ \sum_{e\in\mathcal{I}_d^{(b)}}
\alpha_e^{(b)}
\mathcal{N}_{h_d(t)}\!\left(p_e(t)\right),
\qquad
\mathcal{N}_{h}(u)=
\frac{\lVert h\rVert_2}{\lVert u\rVert_2+\varepsilon}u .
\label{eq:delta_write}
\end{equation}
Since $\mathcal{N}_{h_d}$ matches the payload norm to that of the target
state, the write is determined by both its strength and the direction of the
carried payload. The specific normalization protocols are in App.~\ref{app:preln_repair}.
\subsection{Learning to Route}

Once the payload is fixed, routing amounts to selecting a weighted subgraph of
$\mathcal{G}_C$. For each block $b$, every candidate edge receives a binary
gate and a nonnegative amplitude,
\begin{equation}
m_e^{(b)}\in\{0,1\},
\qquad
a_e^{(b)}\ge0,
\qquad
\alpha_e^{(b)}=m_e^{(b)}a_e^{(b)} .
\label{eq:gated_strength}
\end{equation}
The gate specifies the support of the route, and the amplitude specifies the
write strength on each selected edge. After thresholding, the active graph and
its realized width are
\begin{equation}
\mathcal{E}^{(b)}
=
\{e\in\mathcal{C}:m_e^{(b)}=1\},
\qquad
\kappa_b=|\mathcal{E}^{(b)}|.
\label{eq:active_graph}
\end{equation}

For each block, the router uses a layer-normalized, mean-pooled residual state
from the preceding block as its causal context summary $q_b$. A lightweight
router $g_\phi$ produces gate and
amplitude logits,
\begin{equation}
(\eta^{(b)},\rho^{(b)})=g_\phi(q_b),
\qquad
a^{(b)}=\mathrm{softplus}(\rho^{(b)}).
\label{eq:routing_logits}
\end{equation}
The discrete support is optimized through a Gumbel-Sigmoid relaxation. During
training, Eq.~\ref{eq:delta_write} uses the relaxed strengths
\begin{equation}
\tilde m^{(b)}=\operatorname{GumbelSigmoid}(\eta^{(b)},\tau),
\qquad
\tilde\alpha_e^{(b)}=\tilde m_e^{(b)}a_e^{(b)},
\label{eq:gumbel_route}
\end{equation}
in place of $\alpha_e^{(b)}$. At deployment, the gate is thresholded,
\begin{equation}
m_e^{(b)}
=
\mathbbm{1}\{\sigma(\eta_e^{(b)})>\xi\},
\qquad
\alpha_e^{(b)}=m_e^{(b)}a_e^{(b)} .
\label{eq:threshold_route}
\end{equation}
We write $\operatorname{Route}_\phi(q_b)$ for this gate--amplitude mapping. The base Transformer remains frozen and only the router parameters $\phi$ are
optimized. For a sequence $x=(x_1,\ldots,x_T)$, let $p_{\theta,\phi}$ denote the
predictive distribution induced by the feedback-augmented forward pass, with
$\theta$ fixed. We train the router with next-token language-model loss and
lightweight route regularization:
\begin{equation}
\mathcal{L}(\phi)
=
\mathbb{E}_{x}
\left[
-\frac{1}{T-1}\sum_{t=1}^{T-1}
\log p_{\theta,\phi}(x_{t+1}\mid x_{\le t})
\right]
+
\frac{\lambda_m}{|\mathcal{C}|}
\mathbb{E}_{b}\|\tilde m^{(b)}\|_1
+
\frac{\lambda_\alpha}{|\mathcal{C}|}
\mathbb{E}_{b}\|\tilde\alpha^{(b)}\|_2^2 .
\label{eq:router_loss}
\end{equation}
The language-model term scores each route by its effect on prediction, while
the auxiliary terms favor selective supports and bounded write strengths.
Here \(\tilde m^{(b)}\) and \(\tilde\alpha^{(b)}\) denote vectors over all
candidate edges. We train the router under synchronous execution and reuse the
same router for both synchronous and streaming evaluation.

\subsection{Execution Schedules}
\label{sec:execution-schedules}

The same write operator admits two execution schedules, corresponding to two
ways of reusing deeper computation. The \textbf{synchronous} schedule performs
retrospective refinement: the model first completes a traversal and then uses
each token's deeper computation to take a second look at its earlier
representation. The \textbf{streaming} schedule instead makes refinement
continuous, carrying the deeper interpretation formed by the processed prefix
forward to shape the next token's representation.

\textbf{Synchronous schedule.}
The first traversal records no-feedback reference states
$z^{\mathrm{ref}}=z^{(0)}$ and forms the current-token
payload,
\begin{equation}
p_e^{\mathrm{sync}}(t)
=
p_e(t)
=
z_{s_e}^{(0)}(t)-z_{d_e}^{(0)}(t).
\label{eq:synchronous_payload}
\end{equation}
The model then performs a second traversal using
$p_e(t)=p_e^{\mathrm{sync}}(t)$.

\textbf{Streaming schedule.}
The streaming schedule carries a lightweight state across token positions.
Let $\hat z_\ell(t)$ denote the states recorded during its single,
feedback-augmented traversal. At position $t$, the write uses only the cached
payload from the preceding position,
{\
\begin{equation}
p_e^{\mathrm{stream}}(t)
=
P_e(t-1),
\qquad P_e(0)=0,
\label{eq:stream_payload}
\end{equation}
}while the traversal records current states and updates the cache for the next
position with
{\scriptsize
$P_e(t)\leftarrow \hat z_{s_e}(t)-\hat z_{d_e}(t)$.
}
Algorithm~\ref{alg:reflux} in Appendix~\ref{app:reflux_algorithm} summarizes route selection and both execution modes.

\section{Experiments}
\begin{table*}[t]
\centering
\caption{Perplexity on the ten language-modeling corpora (1024-token chunks). Lower is better. We highlight the
\hlfirst{best} and \hlsecond{second-best} results.}
\label{tab:main-ppl}
\scriptsize
\setlength{\tabcolsep}{3pt}
\renewcommand{\arraystretch}{1.1}
\resizebox{\textwidth}{!}{%
\begin{tabular}{@{}l*{25}{c}@{}}
\toprule
\textbf{Corpus}
& \multicolumn{5}{c}{\modelname{Gemma3-1B}}
& \multicolumn{5}{c}{\modelname{Gemma3-4B}}
& \multicolumn{5}{c}{\modelname{Gemma3-12B}}
& \multicolumn{5}{c}{\modelname{Qwen3-4B}}
& \multicolumn{5}{c}{\modelname{Qwen3-8B}} \\
\cmidrule(lr){2-6}
\cmidrule(lr){7-11}
\cmidrule(lr){12-16}
\cmidrule(lr){17-21}
\cmidrule(lr){22-26}
& \textbf{Base} & \textbf{Orig.} & \textbf{TF-Loop} & \textbf{Stream} & \textbf{Sync.}
& \textbf{Base} & \textbf{Orig.} & \textbf{TF-Loop} & \textbf{Stream} & \textbf{Sync.}
& \textbf{Base} & \textbf{Orig.} & \textbf{TF-Loop} & \textbf{Stream} & \textbf{Sync.}
& \textbf{Base} & \textbf{Orig.} & \textbf{TF-Loop} & \textbf{Stream} & \textbf{Sync.}
& \textbf{Base} & \textbf{Orig.} & \textbf{TF-Loop} & \textbf{Stream} & \textbf{Sync.} \\
\midrule
arXiv
& 26.7 & 23.7 & 26.5 & \hlsecond{22.9} & \hlfirst{\textbf{22.5}}
& 22.3 & 19.2 & 22.0 & \hlsecond{18.4} & \hlfirst{\textbf{18.0}}
& 36.9 & 26.3 & 35.9 & \hlsecond{24.7} & \hlfirst{\textbf{23.7}}
& 18.3 & 16.5 & 18.2 & \hlsecond{15.7} & \hlfirst{\textbf{15.4}}
& 14.5 & 12.8 & 14.3 & \hlsecond{12.3} & \hlfirst{\textbf{12.1}} \\
BigPatent
& 11.8 & 10.7 & 12.0 & \hlsecond{10.6} & \hlfirst{\textbf{10.4}}
& 9.7 & 9.0 & 9.6 & \hlsecond{8.7} & \hlfirst{\textbf{8.5}}
& 20.3 & 15.4 & 20.0 & \hlsecond{14.5} & \hlfirst{\textbf{14.0}}
& 7.8 & 7.3 & 7.9 & \hlsecond{7.1} & \hlfirst{\textbf{7.0}}
& 6.6 & 6.2 & 6.6 & \hlsecond{6.0} & \hlfirst{\textbf{5.9}} \\
BillSum
& 5.2 & 5.1 & 5.1 & \hlfirst{\textbf{4.9}} & \hlsecond{5.0}
& 4.0 & 3.8 & 4.0 & \hlsecond{3.7} & \hlfirst{\textbf{3.6}}
& 4.8 & 4.3 & 4.9 & \hlsecond{4.1} & \hlfirst{\textbf{4.0}}
& 4.7 & \hlsecond{4.5} & 4.6 & 4.6 & \hlfirst{\textbf{4.3}}
& 3.9 & 3.7 & 4.0 & \hlsecond{3.6} & \hlfirst{\textbf{3.5}} \\
BookSum/Book
& 30.6 & 25.6 & 30.8 & \hlsecond{25.0} & \hlfirst{\textbf{24.4}}
& 29.7 & 25.0 & 29.1 & \hlsecond{23.9} & \hlfirst{\textbf{23.3}}
& 78.2 & 53.8 & 75.3 & \hlsecond{50.0} & \hlfirst{\textbf{47.2}}
& 17.9 & 16.3 & 17.7 & \hlsecond{15.8} & \hlfirst{\textbf{15.5}}
& 12.4 & 11.4 & 12.2 & \hlsecond{10.9} & \hlfirst{\textbf{10.8}} \\
C4/WebTextLike
& 25.4 & 24.1 & 25.2 & \hlsecond{23.5} & \hlfirst{\textbf{23.3}}
& 23.8 & 22.0 & 23.5 & \hlsecond{21.2} & \hlfirst{\textbf{20.9}}
& 40.6 & 37.8 & 40.1 & \hlsecond{35.4} & \hlfirst{\textbf{34.8}}
& 21.6 & 19.9 & 21.8 & \hlsecond{19.0} & \hlfirst{\textbf{18.7}}
& 16.3 & 15.6 & 16.2 & \hlsecond{14.7} & \hlfirst{\textbf{14.5}} \\
GovReport
& 11.9 & 10.9 & 12.2 & \hlsecond{10.6} & \hlfirst{\textbf{10.4}}
& 10.9 & 9.9 & 11.0 & \hlsecond{9.4} & \hlfirst{\textbf{9.3}}
& 28.1 & 19.5 & 28.5 & \hlsecond{18.5} & \hlfirst{\textbf{17.5}}
& 9.9 & 9.5 & 9.8 & \hlsecond{9.0} & \hlfirst{\textbf{8.9}}
& 7.9 & 7.6 & 8.0 & \hlsecond{7.3} & \hlfirst{\textbf{7.2}} \\
LAMBADA
& 36.7 & \hlsecond{35.9} & 36.1 & 36.0 & \hlfirst{\textbf{34.8}}
& 28.2 & 26.5 & 27.7 & \hlsecond{25.4} & \hlfirst{\textbf{25.1}}
& 25.8 & 24.8 & 25.3 & \hlsecond{23.6} & \hlfirst{\textbf{23.3}}
& 35.5 & 31.9 & 34.9 & \hlsecond{30.5} & \hlfirst{\textbf{30.0}}
& 27.8 & 25.4 & 27.4 & \hlsecond{24.7} & \hlfirst{\textbf{24.3}} \\
Newsroom
& 14.3 & 13.9 & 14.5 & \hlsecond{13.6} & \hlfirst{\textbf{13.4}}
& 11.7 & 11.2 & 11.9 & \hlfirst{\textbf{10.6}} & \hlsecond{10.7}
& 13.7 & 12.4 & 14.0 & \hlsecond{11.7} & \hlfirst{\textbf{11.4}}
& 20.6 & 18.7 & 20.9 & \hlsecond{18.0} & \hlfirst{\textbf{17.7}}
& 15.2 & 13.9 & 15.3 & \hlsecond{13.5} & \hlfirst{\textbf{13.3}} \\
PG19
& 22.4 & 19.0 & 22.2 & \hlsecond{18.3} & \hlfirst{\textbf{17.8}}
& 19.5 & 16.4 & 19.3 & \hlsecond{15.7} & \hlfirst{\textbf{15.3}}
& 53.1 & 35.4 & 51.5 & \hlsecond{33.3} & \hlfirst{\textbf{31.3}}
& 25.8 & 22.8 & 25.6 & \hlsecond{21.7} & \hlfirst{\textbf{21.2}}
& 19.0 & 17.3 & 18.9 & \hlsecond{16.3} & \hlfirst{\textbf{16.1}} \\
PubMed
& 16.2 & 14.4 & 16.3 & \hlsecond{14.0} & \hlfirst{\textbf{13.8}}
& 11.8 & 10.7 & 11.6 & \hlsecond{10.3} & \hlfirst{\textbf{10.1}}
& 26.1 & 19.6 & 25.6 & \hlsecond{18.3} & \hlfirst{\textbf{17.5}}
& 11.6 & 10.8 & 11.5 & \hlsecond{10.5} & \hlfirst{\textbf{10.4}}
& 9.1 & 8.7 & 9.0 & \hlfirst{\textbf{8.3}} & \hlsecond{8.4} \\
\midrule
\textbf{Mean}
& 20.1 & 18.3 & 20.1 & \hlsecond{17.9} & \hlfirst{\textbf{17.6}}
& 17.2 & 15.4 & 17.0 & \hlsecond{14.7} & \hlfirst{\textbf{14.5}}
& 32.8 & 24.9 & 32.1 & \hlsecond{23.4} & \hlfirst{\textbf{22.5}}
& 17.4 & 15.8 & 17.3 & \hlsecond{15.2} & \hlfirst{\textbf{14.9}}
& 13.3 & 12.3 & 13.2 & \hlsecond{11.8} & \hlfirst{\textbf{11.6}} \\
\bottomrule
\end{tabular}
}
\end{table*}

\begin{table*}[t]
\centering
\caption{Accuracy on eight reasoning benchmarks. Each model group reports
the base model, Original Recirculation, Training-Free Looped,
ReFlux-streaming, and ReFlux-synchronous. Higher is better. We highlight the
\hlfirst{best} and \hlsecond{second-best} results.}
\label{tab:main-accuracy}
\tiny
\renewcommand{\arraystretch}{1.03}
\setlength{\tabcolsep}{1.1pt}
\resizebox{\textwidth}{!}{%
\begin{tabular}{@{}l*{15}{c}@{}}
\toprule
\textbf{Benchmark}
& \multicolumn{5}{c}{\modelname{Gemma3-1B}}
& \multicolumn{5}{c}{\modelname{Gemma3-4B}}
& \multicolumn{5}{c}{\modelname{Qwen3-8B}} \\
\cmidrule(lr){2-6}
\cmidrule(lr){7-11}
\cmidrule(lr){12-16}
& \textbf{Base} & \textbf{Orig.} & \textbf{TF-Loop} & \textbf{Stream} & \textbf{Synchronous}
& \textbf{Base} & \textbf{Orig.} & \textbf{TF-Loop} & \textbf{Stream} & \textbf{Synchronous}
& \textbf{Base} & \textbf{Orig.} & \textbf{TF-Loop} & \textbf{Stream} & \textbf{Synchronous} \\
\midrule

ARC-Easy
& 73.2 & 73.4\accgood{0.2} & 74.5\accgood{1.3} & \hlfirst{\textbf{74.8}\accgood{1.6}} & \hlsecond{74.7\accgood{1.5}}
& 82.9 & 83.0\accgood{0.1} & 83.7\accgood{0.8} & \hlsecond{84.6\accgood{1.7}} & \hlfirst{\textbf{84.9}\accgood{2.0}}
& 80.9 & 81.2\accgood{0.3} & 81.6\accgood{0.7} & \hlsecond{82.1\accgood{1.2}} & \hlfirst{\textbf{82.5}\accgood{1.6}} \\

ARC-Challenge
& 35.7 & 35.8\accgood{0.1} & 36.5\accgood{0.8} & \hlsecond{37.4\accgood{1.7}} & \hlfirst{\textbf{37.8}\accgood{2.1}}
& 56.2 & 56.4\accgood{0.2} & 57.5\accgood{1.3} & \hlsecond{58.0\accgood{1.8}} & \hlfirst{\textbf{58.4}\accgood{2.2}}
& 56.7 & 56.7\accgood{0.0} & 57.4\accgood{0.7} & \hlsecond{58.0\accgood{1.3}} & \hlfirst{\textbf{58.5}\accgood{1.8}} \\

MMLU
& 26.2 & 26.3\accgood{0.1} & \hlfirst{\textbf{27.8}\accgood{1.6}} & \hlsecond{27.5\accgood{1.3}} & 27.4\accgood{1.2}
& 59.5 & 59.9\accgood{0.4} & \hlsecond{60.7\accgood{1.2}} & 60.6\accgood{1.1} & \hlfirst{\textbf{60.8}\accgood{1.3}}
& 72.9 & 73.1\accgood{0.2} & \hlfirst{\textbf{74.3}\accgood{1.4}} & \hlsecond{74.1\accgood{1.2}} & 73.9\accgood{1.0} \\

HellaSwag
& 62.1 & 62.0\accbad{0.1} & 62.0\accbad{0.1} & \hlfirst{\textbf{62.6}\accgood{0.5}} & \hlsecond{62.5\accgood{0.4}}
& 77.2 & 77.0\accbad{0.2} & 77.0\accbad{0.2} & \hlsecond{77.6\accgood{0.4}} & \hlfirst{\textbf{77.9}\accgood{0.7}}
& 74.9 & 74.8\accbad{0.1} & 75.0\accgood{0.1} & \hlsecond{75.3\accgood{0.4}} & \hlfirst{\textbf{75.5}\accgood{0.6}} \\

Winogrande
& 58.3 & 58.2\accbad{0.1} & 58.4\accgood{0.1} & \hlsecond{58.6\accgood{0.3}} & \hlfirst{\textbf{58.7}\accgood{0.4}}
& 64.4 & 64.4\accgood{0.0} & 64.6\accgood{0.2} & \hlfirst{\textbf{64.9}\accgood{0.5}} & \hlsecond{64.8\accgood{0.4}}
& 67.7 & 67.8\accgood{0.1} & 67.4\accbad{0.3} & \hlsecond{68.0\accgood{0.3}} & \hlfirst{\textbf{68.2}\accgood{0.5}} \\

GSM8K
& 1.7 & 1.9\accgood{0.2} & 2.1\accgood{0.4} & \hlsecond{3.7\accgood{2.0}} & \hlfirst{\textbf{4.9}\accgood{3.2}}
& 38.4 & 38.5\accgood{0.1} & 39.1\accgood{0.7} & \hlsecond{41.2\accgood{2.8}} & \hlfirst{\textbf{41.5}\accgood{3.1}}
& 88.6 & 88.9\accgood{0.3} & 88.8\accgood{0.2} & \hlsecond{90.1\accgood{1.5}} & \hlfirst{\textbf{90.4}\accgood{1.8}} \\

HotpotQA
& 42.3 & 42.4\accgood{0.1} & 42.7\accgood{0.4} & \hlsecond{45.5\accgood{3.2}} & \hlfirst{\textbf{47.0}\accgood{4.7}}
& 49.4 & 49.6\accgood{0.2} & 49.2\accbad{0.2} & \hlsecond{53.3\accgood{3.9}} & \hlfirst{\textbf{53.5}\accgood{4.1}}
& 57.2 & 57.3\accgood{0.1} & 57.6\accgood{0.4} & \hlsecond{61.4\accgood{4.2}} & \hlfirst{\textbf{61.7}\accgood{4.5}} \\

2WikiMultiHopQA
& 24.6 & 24.7\accgood{0.1} & 24.9\accgood{0.3} & \hlsecond{28.3\accgood{3.7}} & \hlfirst{\textbf{29.7}\accgood{5.1}}
& 38.5 & 38.6\accgood{0.1} & 38.9\accgood{0.4} & \hlsecond{42.8\accgood{4.3}} & \hlfirst{\textbf{43.2}\accgood{4.7}}
& 44.5 & 44.7\accgood{0.2} & 45.0\accgood{0.5} & \hlsecond{49.1\accgood{4.6}} & \hlfirst{\textbf{49.5}\accgood{5.0}} \\

\midrule
\textbf{Mean}
& \textbf{40.5} & \textbf{40.6}\accgood{0.1} & \textbf{41.1}\accgood{0.6} & \hlsecond{\textbf{42.3}\accgood{1.8}} & \hlfirst{\textbf{42.8}\accgood{2.3}}
& \textbf{58.3} & \textbf{58.4}\accgood{0.1} & \textbf{58.8}\accgood{0.5} & \hlsecond{\textbf{60.4}\accgood{2.1}} & \hlfirst{\textbf{60.6}\accgood{2.3}}
& \textbf{67.9} & \textbf{68.1}\accgood{0.2} & \textbf{68.4}\accgood{0.5} & \hlsecond{\textbf{69.8}\accgood{1.9}} & \hlfirst{\textbf{70.0}\accgood{2.1}} \\
\bottomrule
\end{tabular}
}
\end{table*}

\begin{figure}[t]
    \centering
    \begin{minipage}[c]{0.382\linewidth}
        \small
        \captionof{figure}{Deep computation as a predictive signal.
        In \emph{``The \textcolor{TrophyColor}{trophy} does not fit in the \textcolor{SuitcaseColor}{suitcase} because \textcolor{PronounColor}{it} is too big,''} shallow layers may encode only coarse semantics when processing \textcolor{PronounColor}{\textit{``it''}}, while previous deeper computation can establish the relation between the \textcolor{TrophyColor}{\textit{trophy}} and \textcolor{SuitcaseColor}{\textit{suitcase}}. Instead of revisiting the same token to apply this refinement, ReFlux-streaming carries the $\Delta$ forward and makes it available to subsequent tokens, allowing them to benefit from computation already performed.}
        \label{fig:case1}
    \end{minipage}
    \hfill
    \begin{minipage}[c]{0.61\linewidth}
        \centering
        \includegraphics[width=\linewidth]{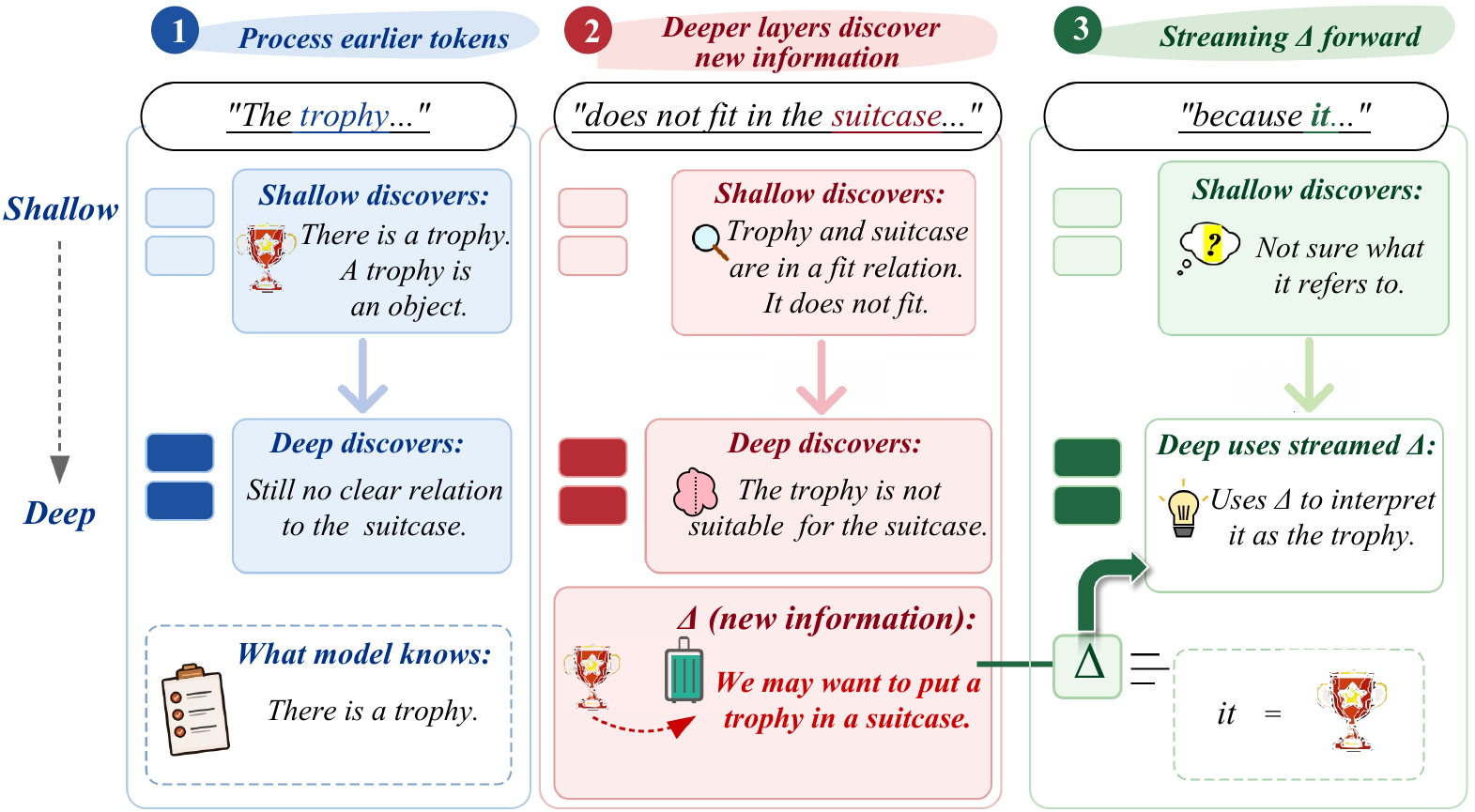}
    \end{minipage}
\end{figure}
\begin{table*}[t]
\centering
\caption{Decoding efficiency and C4 language-modeling quality. Efficiency metrics are color-coded.}
\label{tab:efficiency}
\scriptsize
\setlength{\tabcolsep}{4.5pt}
\renewcommand{\arraystretch}{0.95}
\resizebox{\textwidth}{!}{%
\begin{tabular}{@{}lccccc|ccccc@{}}
\toprule
& \multicolumn{5}{c|}{\modelname{Gemma3-1B}}
& \multicolumn{5}{c}{\modelname{Qwen3-4B}} \\
\cmidrule(lr){2-6} \cmidrule(l){7-11}
\textbf{Method}
& \textbf{Runtime($\times$)}
& \textbf{FLOPs($\times$)}
& \textbf{Memory(GB)}
& \textbf{Latency(ms/tok)}
& \textbf{PPL}
& \textbf{Runtime($\times$)}
& \textbf{FLOPs($\times$)}
& \textbf{Memory(GB)}
& \textbf{Latency(ms/tok)}
& \textbf{PPL} \\
\midrule
Base
& \csA{1.00} & \csA{1.00} & 1.88 & 15.05 & 25.4
& \csA{1.00} & \csA{1.00} & 7.56 & 16.05 & 21.6 \\
Orig.
& \csH{1.88} & \csL{2.00} & 1.90 & 28.26 & 24.1
& \csJ{1.94} & \csL{2.00} & 7.58 & 31.14 & 19.9 \\
TF-Loop
& \csG{1.34} & \csF{1.31} & 1.88 & 20.16 & 25.2
& \csD{1.21} & \csE{1.22} & 7.56 & 19.42 & 21.8 \\
Streaming
& \csB{1.02} & \csA{1.00} & 1.88 & 15.34 & 23.5
& \csC{1.08} & \csA{1.00} & 7.57 & 17.35 & 19.0 \\
Synchronous
& \csI{1.91} & \csL{2.00} & 1.89 & 28.71 & 23.3
& \csK{1.97} & \csL{2.00} & 7.64 & 31.59 & 18.7 \\
\bottomrule
\end{tabular}}
\end{table*}



\label{sec:ex}
\vspace{-0.4em}
In this section, we conduct extensive experiments to answer: Does ReFlux improve performance across models (Sec \ref{sec:main-results})? 
How do its gains trade off against
computational cost (Sec \ref{sec:efficiency})?  What
mechanisms underlie its effectiveness (Sec \ref{sec:understanding-reflux})? 

\subsection{Experiment Setup}

\label{sec:exp-setup}
\paragraph{Models and Data.}
We evaluate ReFlux on five models:
Gemma3-1B, Gemma3-4B, Gemma3-12B, Qwen3-4B, and Qwen3-8B \citep{team2025gemma,yang2025qwen3}.
The router is trained on a compact calibration mixture from the training
splits of C4 \citep{raffel2020exploring}, arXiv \citep{gao2020pile}, and
PG19 \citep{rae2019compressive}, covering general web text, scientific
writing, and long-form documents. We use 4,096 1,024-token windows from each
calibration corpus for router training. For language modeling, we use ten corpora: arXiv \citep{gao2020pile}, BigPatent \citep{sharma2019bigpatent}, BillSum \citep{kornilova2019billsum}, BookSum \citep{kryscinski2022booksum}, C4 \citep{raffel2020exploring}, GovReport \citep{huang2021efficient},
LAMBADA \citep{paperno2016lambada}, Newsroom \citep{grusky2018newsroom}, PG19 \citep{rae2019compressive}, and PubMed \citep{gao2020pile}. We use the validation split for C4 and
PG19 and the test split for the remaining corpora. We evaluate reasoning with eight standard benchmarks: MMLU \citep{hendrycks2020measuring}, ARC-Easy,
ARC-Challenge \citep{clark2018think}, HellaSwag \citep{zellers2019hellaswag}, Winogrande \citep{sakaguchi2021winogrande}, GSM8K \citep{cobbe2021training}, HotpotQA \citep{yang2018hotpotqa}, and
2WikiMultiHopQA \citep{ho2020constructing}.
HotpotQA and 2WikiMultiHopQA provide complementary multi-hop settings in
which the required evidence is distributed across passages, while the other
benchmarks measure knowledge, science, and 
reasoning under shorter contexts.

\paragraph{Baselines.}
We compare against the base model, the original
Recirculation \citep{mozer2026recirculation}, and Training-Free Looped
Transformers (TF-Loop) \citep{chen2026training}. TF-Loop applies a contiguous intermediate
layer window to the model with damped refinement.

\paragraph{Implementation Details.}
We partition each input
into routing blocks and make one causal routing decision per block. The router consists of a
two-layer MLP with hidden width 512 and GELU nonlinearity, followed by separate
linear heads for edge gates and write amplitudes. We set $\xi=0.6$, $\lambda_m=10^{-3}$ and
$\lambda_\alpha=10^{-2}$.  
Full details are given
in App.~\ref{app:details}.

\subsection{Main Results}
\label{sec:main-results}

\begin{figure*}[t]
\centering
\includegraphics[width=0.98\textwidth]{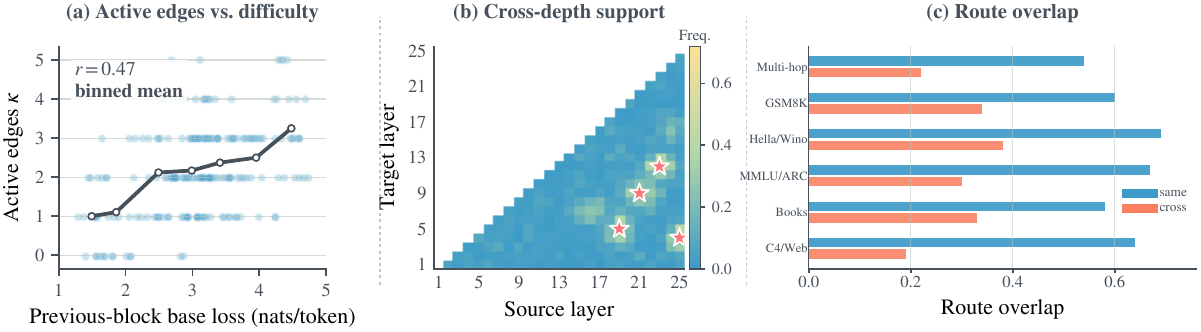}
\caption{Routing behavior summary. (a) Integer active-edge count for routing
block $b{+}1$ against the base-model NLL of the preceding block $b$; the
line connects binned means. (b) Source-target heatmap of aggregated activation
frequencies, with the most frequently used pairs marked by stars. (c)
Comparison of same-family and cross-family route overlap.}

\label{fig:routing-behavior}
\end{figure*}

Tables~\ref{tab:main-ppl} and~\ref{tab:main-accuracy} show that ReFlux improves both language modeling and reasoning across model families.
On the ten-corpus PPL suite, ReFlux consistently outperforms both the Base model and Original Recirculation across all corpora. The improvement remains consistent across the Gemma3 and Qwen3 families and across different model scales. Original Recirculation also helps but remains consistently weaker,
while TF-Loop often stays close to the base model. On downstream benchmarks,
ReFlux improves average accuracy from $40.5$ to $42.3$/$42.8$ on Gemma3-1B,
from $58.3$ to $60.4$/$60.6$ on Gemma3-4B, and from $67.9$ to
$69.8$/$70.0$ on Qwen3-8B. The gains are largest on HotpotQA and
2WikiMultiHopQA, where answers require combining evidence across passages.
We also compare ReFlux with CoT \citep{wei2022chain} and their
combination in App.~\ref{app:cot}.
Together, the results suggest that the benefit is not merely re-executing
depth, but accessing the computation that was newly produced.

\subsection{Efficiency Analysis}
\label{sec:eff}
\label{sec:efficiency}

 Table~\ref{tab:efficiency} compares these gains against inference cost.
  ReFlux-streaming keeps the backbone at \(1\times\) FLOPs, adds almost no
  memory, and runs at only \(1.02\times\)/\(1.08\times\) wall-clock cost during
  decoding, yet its PPL remains close to synchronous execution. As demonstrated in
  Fig.~\ref{fig:case1}, useful feedback need not always revisit the same token:
  the newly accumulated \(\Delta\) can  be carried forward and bias the interpretation of subsequent
  tokens, serving as a predictive signal. Synchronous ReFlux gives the strongest PPL by performing a stricter
  second traversal, but its roughly \(1.9\times\) runtime matches the cost profile
  of Original Recirculation. Details are given in App.~\ref{app:efficiency_protocol}, with a Pareto view in Fig.~\ref{fig:eff_pareto}.

\subsection{Understanding ReFlux}

\begin{wraptable}{r}{0.38\linewidth}
\centering
\vspace{-3em}
\caption{Ablations on Gemma3-4B. We report C4 PPL and two reasoning accuracies, ARC-E and ARC-C.}
\label{tab:reflux-ablation}
\scriptsize
\begin{adjustbox}{width=\linewidth}
\begin{tabular}{@{}lccc@{}}
\toprule
Config & C4 PPL $\downarrow$ & ARC-E $\uparrow$ & ARC-C $\uparrow$ \\
\midrule
ReFlux-Sync. & \textbf{20.9} & \textbf{84.9} & \textbf{58.4} \\
Base & 23.8\bad{2.9} & 82.9\accbad{2.0} & 56.2\accbad{2.2} \\
Learned-full & 22.1\bad{1.2} & 83.1\accbad{1.8} & 55.9\accbad{2.5} \\
Uniform-strength & 21.6\bad{0.7} & 84.1\accbad{0.8} & 57.3\accbad{1.1} \\
Fixed-Delta & 21.7\bad{0.8} & 84.2\accbad{0.7} & 57.5\accbad{0.9} \\
Heuristic-Delta & 24.3\bad{3.4} & 82.0\accbad{2.9} & 55.8\accbad{2.6} \\
\bottomrule
\end{tabular}
\end{adjustbox}
\end{wraptable}

\label{sec:understanding-reflux}
\paragraph{Routing Behavior.}

We  ask whether the router learns a genuinely input-conditioned policy. For each
block, Fig.\ref{fig:routing-behavior} shows three consistent routing patterns: higher-loss context leads the subsequent route to activate more edges, frequently used source-target pairs form localized cross-depth hotspots, and routes are more stable within task families than across families. These results suggest that the router learns a structured, input-conditioned computation policy. Details are given in
Appendix~\ref{app:routing_details}.

\paragraph{Ablations.}
Table~\ref{tab:reflux-ablation} summarizes the ablations, with the full protocol in App.~\ref{app:ablation_protocol}. Replacing the increment with the full source state degrades performance, showing that the depth increment is a more effective feedback payload. Removing either strength adaptation or input-conditioned support also causes consistent degradation, indicating that both write strength and route selection contribute. In contrast, Heuristic-Delta performs substantially worse and is the only variant to fall below the base model, suggesting that static representation geometry alone is insufficient to determine which increments are useful for each input.
We further compares ReFlux with trainable controls in App.~\ref{app:budget_matched}.

\begin{figure}[t]
\centering
\includegraphics[width=0.98\columnwidth]{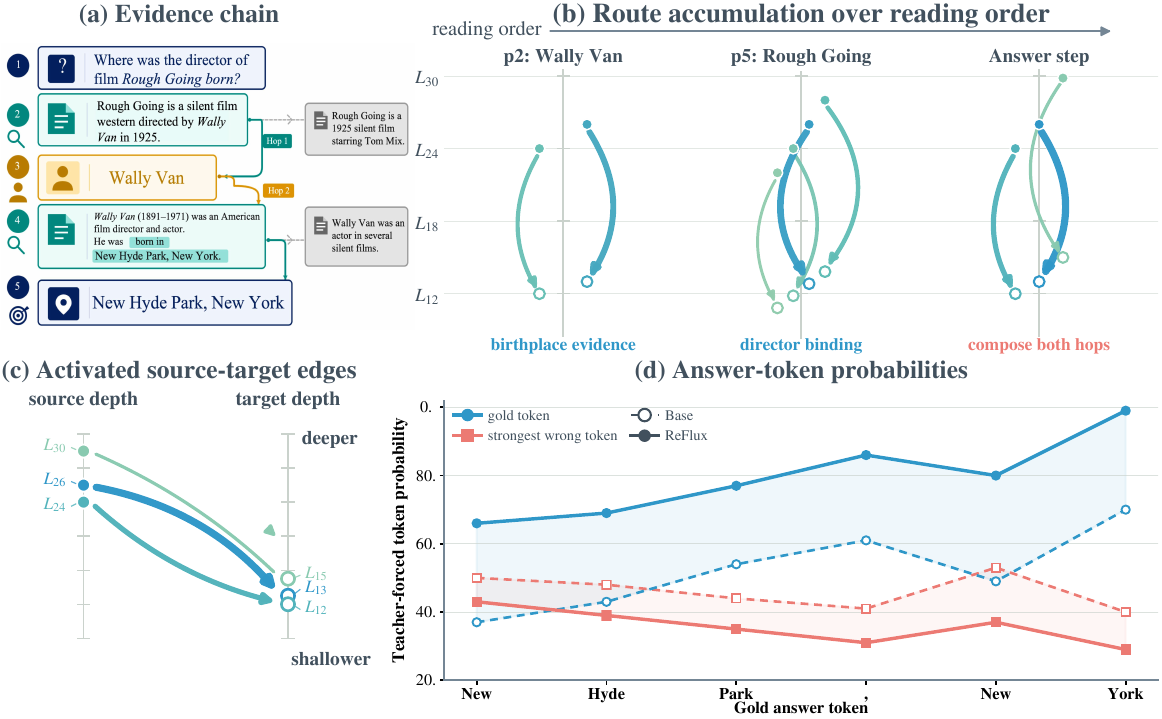}
\caption{Case study (App.~\ref{app:complete_case}). (a) The question's evidence chain, (b) how route support accumulates (we set the block size to 1 token for visualization), (c) the final active source-target
edges, and (d) teacher-forced token probabilities for the gold answer path and
the strongest wrong competitor.}
\label{fig:reflux-case-study}
\end{figure}

\paragraph{Case Study.}
We use one representative 2WikiMultiHopQA example to illustrate how ReFlux
supports evidence composition across hops and states (Fig.\ref{fig:reflux-case-study}).
The question requires first binding \emph{Rough Going} to \emph{Wally Van} and
then carrying that bridge entity to the birthplace answer.  ReFlux concentrates
its routed support on the bridge passage, so the final decision reuses the
earlier hop rather than rebuilding it from scratch. In the trace, this leads to a larger gap between the gold answer path and the strongest distractor, enabling ReFlux to answer correctly where the base model fails.

\section{Conclusion}

 In this work, we identified the depth increment ($\Delta$) as a precise, composable feedback signal in Transformers and proposed ReFlux, a learnable, input-conditioned feedback graph over frozen models. Extensive experiments across various models demonstrated its effectiveness. Looking forward, integrating this adaptive $\Delta$-routing into pre-training or test-time compute presents exciting directions. More broadly, ReFlux shows that frozen LLMs can recover useful capability by letting later computation reshape earlier
 interpretation, giving the model a second look with new discoveries.

\section{AI Usage Statement}
\label{app:llm_usage}

During the preparation of this manuscript, we used advanced language models solely for editorial assistance. Their use was restricted to language-level tasks, including grammar correction, sentence refinement, and improving the clarity and coherence of the presentation. The models were not used to generate or determine the research ideas, technical methodology, experimental design, or reported findings. All scientific contributions and conclusions presented in this work were developed and verified by the authors.
\section{Ethics Statement}
This work does not involve human subjects, personally identifiable information, or the collection of sensitive or private data. All experiments are conducted on publicly available pretrained models and datasets, following their respective usage and licensing terms. The proposed method is intended to study efficient and controllable information propagation in language models and does not introduce any new data collection or human-facing deployment. We are not aware of any specific ethical, privacy, security, or fairness concerns arising from the methodology or experiments presented in this work. We have also disclosed our source code, experimental protocols and evaluation procedures to support transparent and reproducible research.

\section{Reproducibility Statement}
We provide detailed descriptions of the ReFlux architecture, feedback construction,
routing mechanism, and inference procedures in Section~\ref{sec:method} and
Appendix~\ref{app:reflux_algorithm}. The experimental setup, including the
evaluated models, datasets, prompts, evaluation protocols, and implementation
details, is documented in Section~\ref{sec:exp-setup} and Appendix~\ref{app:details}. Code implementation can be found at
\href{https://github.com/gooogleshanghai/reflux}{\texttt{https://github.com/gooogleshanghai/reflux}}.

\bibliography{iclr2027_conference}
\bibliographystyle{iclr2027_conference}

\appendix

\section{Extended Related Works}
\label{app:more_related}
\begin{table*}[t]
\centering
\caption{Positioning of ReFlux among depth-feedback and
activation-editing methods. \textcolor{green!60!black}{\ding{51}} /
\textcolor{red!70!black}{\ding{55}}: property holds / does not hold;
\partc: partially. $\dagger$requires pretraining or continued
training. The single-traversal column for ReFlux refers to its
streaming schedule.
{\footnotesize Sources: UT \citep{dehghani2018universal}; Looped LM
\citep{yang2024looped}; SMELT \citep{wang2026smelt}; MoR
\citep{bae2026mixture}; MoDA \citep{zhu2026moda}; T$^2$MLR
\citep{cai2026t}; TF-Loop
\citep{chen2026training}; CoLa \citep{li2025skip}; CAA
\citep{rimsky2024steering}; ITI \citep{li2023inference}; Patchscopes
\citep{ghandeharioun2024patchscopes}; Recirculation
\citep{mozer2026recirculation}.}}
\label{tab:positioning}
\scriptsize
\setlength{\tabcolsep}{2.5pt}
\renewcommand{\arraystretch}{1.12}

\resizebox{\textwidth}{!}{%
\begin{tabular}{@{}lccccccc@{}}
\toprule
\textbf{Method}
& \textbf{\shortstack{adds effective\\depth}}
& \textbf{\shortstack{no backbone\\retraining}}
& \textbf{\shortstack{any frozen\\checkpoint}}
& \textbf{\shortstack{single\\traversal}}
& \textbf{\shortstack{in-place\\write}}
& \textbf{\shortstack{endogenous\\payload}}
& \textbf{\shortstack{multi-edge\\feedback}} \\
\midrule
UT / Looped LM / SMELT$^\dagger$
& \goodc & \badc & \badc & \badc & \goodc & \goodc & \badc \\
MoR$^\dagger$
& \goodc & \badc & \badc & \partc & \goodc & \goodc & \badc \\
MoDA$^\dagger$
& \partc & \badc & \badc & \goodc & \goodc & \goodc & \partc \\
T$^2$MLR$^\dagger$
& \goodc & \badc & \badc & \goodc & \goodc & \goodc & \badc \\
TF-Loop
& \goodc & \goodc & \goodc & \badc & \goodc & \goodc & \badc \\
CoLa
& \goodc & \goodc & \goodc & \badc & \goodc & \goodc & \partc \\
CAA / ITI
& \badc & \goodc & \goodc & \goodc & \goodc & \badc & \badc \\
Patchscopes
& \badc & \goodc & \goodc & \badc & \badc & \goodc & \badc \\
Recirculation
& \goodc & \goodc & \goodc & \badc & \goodc & \goodc & \badc \\
\rowcolor{E8F1FF}
ReFlux (ours)
& \goodc & \goodc & \goodc & \goodc & \goodc & \goodc & \goodc \\
\bottomrule
\end{tabular}%
}
\end{table*}
\subsection{Residual Streams}

The residual stream provides the basic substrate for the depth-wise information flow considered by ReFlux. Originating from the residual connections of ResNets \citep{he2016deep}, this additive structure preserves an identity path while accumulating learned updates across layers. In Transformers, layers read from and write to a shared residual stream, which can therefore be viewed as a communication channel between components \citep{elhage2021mathematical}. Normalization further shapes how these updates accumulate: pre-norm architectures improve optimization stability in deep Transformers \citep{xiong2020layer}, while subsequent work identifies reduced contributions from later blocks and explores alternatives such as Peri-LN and Mix-LN \citep{kim2025peri,li2024mix}. Recent work has begun to treat this stream not merely as a sequence of overwritten states, but as a structured record of computation across depth. Massive activations and attention sinks show that information flow can become concentrated on particular tokens \citep{sun2024massive}, while Attention Residuals \citep{team2026attention} demonstrates that contributions from different depths can be selectively aggregated. MoDA \citep{zhu2026moda} takes this idea further by giving attention heads explicit access to depth-specific keys and values produced by shallower layers at the same token position. These works suggest that information accumulated along depth can remain individually accessible rather than being represented only through the current full residual state. However, most related designs propagate information forward across depth, from shallower to deeper layers. Two exceptions reverse this direction. T$^2$MLR \citep{cai2026t} trains a gated fusion pathway along time, fusing a cached middle-layer state of the previous token into an early layer of the current token, while Recirculation \citep{mozer2026recirculation} writes full deep states back into shallower streams during inference. ReFlux combines the deep-to-shallow direction of these designs with an increment payload and a frozen backbone: it feeds $\Delta_{d\rightarrow s}=z_s-z_d$, the computation newly accumulated between two depths, back into shallower layers of the same token or, through streaming, of subsequent tokens.

\subsection{Depth Utilization and Redundancy}

A growing body of evidence suggests that depth is used unevenly across layers. ShortGPT \citep{men2025shortgpt} finds that many mid-depth layers can be removed with limited performance loss, while \citet{gromov2025unreasonable} show that some models tolerate removing up to half of their deeper layers; LaCo  \citep{yang2024laco} similarly collapses rear layers into earlier ones with little quality loss. Recent analysis attributes part of this redundancy to normalization: LayerNorm and RMSNorm make block outputs relatively insensitive to input magnitude, causing deeper layers to contribute progressively less \citep{sun2026curse}. Yet reduced contribution does not mean that depth stops computing useful information. Intermediate layers tend to make incremental, directionally faithful updates, whereas later layers can shift predictions toward generic, alignment-preferred tokens, thus decoding from an intermediate entropy-valley layer can outperform decoding from the final layer \citep{zhang2026deeper}. Meanwhile, feed-forward layers act as key--value memories that progressively build predictions by promoting concepts in vocabulary space \citep{geva2021transformer,geva2022transformer}. Causal tracing further localizes factual recall to mid-layer MLPs, where editing a single module can rewrite factual associations \citep{meng2022locating}, while affine probes recover increasingly sharp predictions from intermediate states \citep{belrose2023eliciting}. The picture that emerges is not that deeper computation is redundant, but that its output can be partly redundant even when the computation itself is informative. ReFlux builds on this distinction by propagating the newly computed increment rather than replaying the full state.


\subsection{Cognitive Principles}
Our computation reuse perspective also resonates with a broader view of information processing as hierarchical, recurrent, and selective. Predictive coding proposes that higher cortical areas send predictions to lower areas to refine the interpretation of incoming activity \citep{rao1999predictive}, while the cortical hierarchy is organized through reciprocal connections across levels \citep{Felleman1991DistributedHP}. Cognitive theories further suggest that limited-capacity working memory selectively maintains task-relevant information \citep{Baddeley2003WorkingML}, with processing shaped by the interaction between controlled and automatic mechanisms \citep{shiffrin1977controlled}. Related theories of attention and free energy also emphasize that the precision assigned to a signal can depend on its content \citep{feldman2010attention}. These principles provide a conceptual interpretation of several design choices in ReFlux: hierarchical feedback, selective routing under capacity constraints, content-dependent modulation, and temporally delayed feedback. In particular, ReFlux propagates newly computed depth increments rather than full representations, and its streaming formulation incorporates these corrections into the ongoing feedforward computation, providing a computational analogue of recurrent information exchange in hierarchical processing.

\begin{figure*}[t]
\centering
\includegraphics[width=1\textwidth]{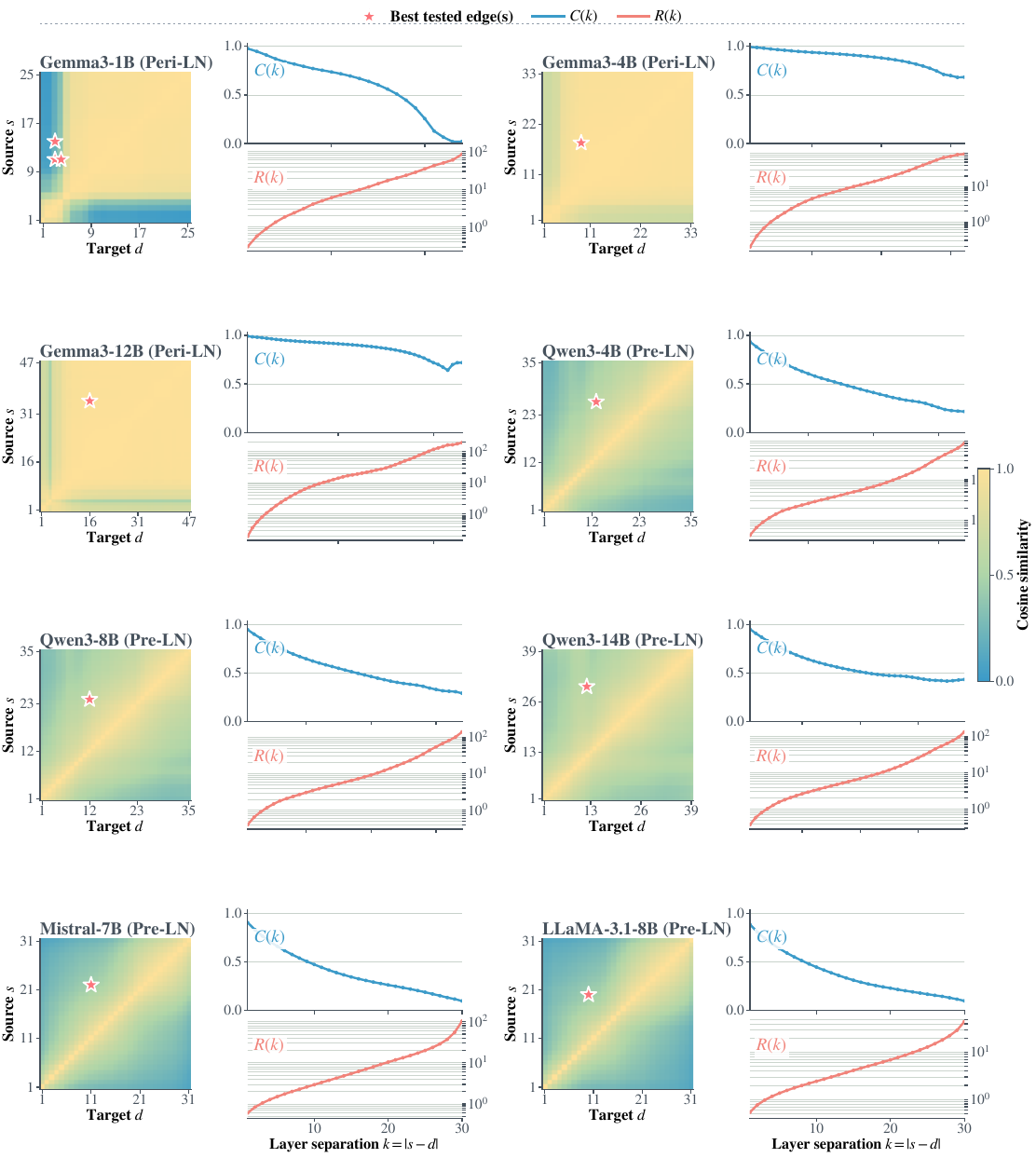}
\caption{Cross-depth residual-stream geometry across the eight profiled
models. Each model panel shows pairwise cosine similarity, cosine similarity
$C(k)$, and norm-relative displacement $R(k)$. Stars denote the best tested source-target edge(s) for each
model.}
\label{fig:cross_depth_all}
\end{figure*}

\section{Diagnosing and Repairing Cross-Model Failure}
\label{app:preln_repair}

The absence of a clear gain from Recirculation on Qwen-Family models has been reported in
prior work. Here we try to explain why the same feedback idea transfers unevenly across normalization
architectures and to test whether the limitation can be removed. The results
below show that the Qwen3 behavior is structured rather than incidental: the
outcome depends jointly on the write strength, the target depth, and the
representation written back.

\subsection{A normalization-dependent failure mode}
Let $n_\ell(t)\coloneqq\lVert z_\ell(t)\rVert_2$ denote the residual-stream
norm trajectory and let
$r_\ell(t)\coloneqq\lVert F_\ell(z_\ell,t)\rVert_2/(n_\ell(t)+\varepsilon)$
denote the relative branch budget at depth $\ell$. In a Pre-LN model, both
quantities are calibrated during training; a Peri-LN model further normalizes
branch outputs before they enter the stream:
\[
z_{\ell+1}^{\mathrm{Peri}}(t)
=
z_\ell(t)+\mathrm{LN}_{\mathrm{out}}\!\bigl(F_\ell(\mathrm{LN}_{\mathrm{in}}(z_\ell(t)))\bigr),
\qquad
z_{\ell+1}^{\mathrm{Pre}}(t)
=
z_\ell(t)+F_\ell(\mathrm{LN}(z_\ell(t))).
\]
Applying a stream-scale feedback write to Qwen3 can therefore introduce a norm
shock
\[
\sigma_d(t)\coloneqq
\frac{\lVert \bar z_d(t)\rVert_2}{\lVert z_d(t)\rVert_2+\varepsilon},
\qquad
\bar z_d(t)=z_d(t)+u_d(t),
\]
where $u_d(t)$ is the injected feedback vector. When $\sigma_d(t)$ departs
substantially from one, the write moves the stream off the calibrated
trajectory used by the downstream stack. This does not merely rescale the representation: the perturbed stream
is still processed by layers whose updates were trained under the original
norm profile, so the shock can persist as a directional error rather than
being washed out. The cross-depth profiles in
App.~\ref{app:cross_depth_geometry} are consistent with this picture: the
residual state is dominated by a low-dimensional direction that rotates only
gradually with depth, so a scale disturbance survives through the remaining
blocks.
\subsection{Our repair}
We retain the increment as the feedback content, but preserve the norm of the
destination stream after writing it. If $\bar z_d$ denotes the stream after
the payload has been added, we use
\begin{equation}
\widetilde z_d(t)
=
\frac{\lVert z_d(t)\rVert_2}
       {\lVert \bar z_d(t)\rVert_2+\varepsilon}
\bar z_d(t).
\label{eq:app_preserve_norm}
\end{equation}
The normalization protocols used in the experiments are architecture-specific.
For Gemma3 (Peri-LN), we use the stream-scale payload normalization
$\mathcal{N}_{z_d}$ in Eq.~\ref{eq:delta_write} without an additional
post-write projection. For Qwen3 (Pre-LN), we use the same target-scale payload
normalization followed by the norm-preserving projection in
Eq.~\ref{eq:app_preserve_norm}. Thus the two models receive the same
increment-based feedback, but under different residual-stream norm protocols.

\subsection{Cross-architecture results}
The repaired operator turns the previously marginal Qwen3 transfer into a
consistent positive result. On Qwen3-4B, it improves C4 perplexity by
$+5.99\%$ at 512 tokens and $+6.15\%$ at 1,024 tokens.

In conclusion, these results turn a previously reported negative transfer
into a more precise architectural account. The limitation is not that
deep-to-shallow feedback is intrinsically ineffective on Pre-LN models, but
that an otherwise informative write can violate the residual-stream scale
under which the downstream computation was trained. Preserving the destination
norm removes this incompatibility while leaving the feedback content itself
unchanged: the useful signal remains the depth increment $\Delta$, and the
normalization protocol determines whether that signal can be safely received.
This distinction separates the semantic design of feedback from its
architecture-dependent execution conditions, and provides the basis for
transferring the method beyond the models studied here through lightweight
stream profiling. 
\section{Cross-depth Geometry Across Models}
\label{app:cross_depth_geometry}

We apply the same residual-stream profiling protocol across eight models from different families on C4.
Fig.\ref{fig:cross_depth_all} provides a compact cross-model view: each
panel contains the pairwise cosine heatmap and the corresponding similarity
and displacement profiles by layer separation. Stars mark the best tested
source-target edge for each model.

\begin{table}[t]
\centering
\caption{Positioning of CoT, filler-token, looped, and increment-feedback
methods. Filler tokens must be trained to be exploited \citep{pfau2024let};
the router calibration of ReFlux is a single lightweight pass. The last column
indicates which of the two bottlenecks each method primarily addresses.
\textcolor{green!60!black}{\ding{51}} /
\textcolor{red!70!black}{\ding{55}}: property holds / does not hold;
\partc: partially.
{\footnotesize Sources: CoT \citep{wei2022chain}; Filler tokens
\citep{pfau2024let}; Looped \citep{dehghani2018universal,yang2024looped}.}}
\label{tab:cot_positioning}
\setlength{\tabcolsep}{3pt}
\renewcommand{\arraystretch}{1.15}
\resizebox{\columnwidth}{!}{%
\begin{tabular}{@{}lccccc@{}}
\toprule
\textbf{Method}
& \textbf{\shortstack{adds serial steps}}
& \textbf{\shortstack{extra training}}
& \textbf{\shortstack{extra token generation}}
& \textbf{\shortstack{plug-and-play}}
& \textbf{\shortstack{bottleneck addressed}} \\
\midrule
CoT
& \goodc & \badc & 5--50$\times$ & \goodc & computation \\
Filler tokens
& \goodc & \goodc & $N\times$ & \badc & computation \\
Looped
& \goodc & \goodc & 0 & \badc & computation \\
\rowcolor{E8F1FF}
ReFlux
& \badc & \partc & 0 & \goodc & \textbf{access} \\
\bottomrule
\end{tabular}}
\end{table}
\section{Comparison with Chain-of-Thought Reasoning}
\label{app:cot}

Since chain-of-thought (CoT) prompting is a standard way of allocating additional
computation on reasoning benchmarks, we further compare ReFlux with CoT from the
perspective of how additional computation is introduced and used.

\subsection{Theoretical Comparison}

In CoT, each generated token provides an additional serial reasoning step,
allowing the model to construct new intermediate states over a longer
computation trajectory. \citet{merrill2024expressive} formalize this view by
characterizing the expressive power of decoder models as a function of their
intermediate computation budget. The role of additional computation is also
highlighted by the filler-token setting of \citet{pfau2024let}, where
meaningless tokens can sometimes match the performance of CoT on controlled
tasks, which suggests that part of the benefit of CoT can arise from the
additional \textit{computation} itself, rather than the semantic content of its
intermediate reasoning steps.

There is an architectural reason why this computation must be purchased in
tokens. In a standard decoder-only model, a deep representation computed at
one position can influence later computation only through two
channels: it is either re-derived by traversing the same depth again and
retrieved indirectly through attention, or compressed into the discrete token
that carries the context forward through the unembed--decode--embed
interface \citep{cai2026t}. Chain-of-thought prompting is best understood as a deliberate
exploitation of the second channel: each reasoning step purchases persistence
for intermediate computation by paying one generated token, so the state that
survives a step boundary is a quantized, low-bandwidth projection of the
richer computation that produced it.

ReFlux introduces a latent feedback channel that preserves intermediate representations without routing them through either additional depth or the discrete token interface.
Rather than increasing the number of serial reasoning steps, it keeps the
depth available to each token fixed while changing how previously computed
representations are reused. In the streaming setting, each token still
undergoes a single depth-$L$ forward pass: an increment computed at an
earlier position is written into a shallow key/value entry and can later be
retrieved directly, without re-executing the layers that produced it or
passing through the token interface. The synchronous schedule realizes the
same feedback through an additional traversal at a fixed $2\times$ FLOPs
cost---a cost that, unlike the token-level computation introduced by
chain-of-thought, does not grow with the length of the reasoning chain.

This distinction separates two different computational bottlenecks. When a
useful intermediate result has not yet been computed, additional serial
computation can create it, as in CoT. When the relevant result has already
been computed but is inaccessible at the depth where it is needed, however,
additional computation is not necessarily the solution. The bottleneck is
then one of \emph{access}: ReFlux makes existing computation available to the
computation that needs it, rather than repeatedly recomputing it through a
longer serial trajectory. Table~\ref{tab:cot_positioning} summarizes this
positioning against CoT, filler-token, and looped approaches.

\subsection{Empirical Comparison}
\label{app:cot_empirical}

We evaluate this analysis on three benchmarks spanning the
computation--access axis: 2WikiMultiHopQA (composition,
access-bound), MMLU (knowledge, neutral), and GSM8K (arithmetic,
computation-bound). On Qwen3-8B, we compare four settings under a unified,
fully generative protocol: direct answering (Base), zero-shot
chain-of-thought prompting (CoT), ReFlux feedback, and their combination
(CoT + ReFlux). In every setting, the model generates the final answer from
scratch; in the CoT setting, it first generates a detailed, step-by-step
reasoning trace before producing the answer. We then extract the answer from
the generated text and report both accuracy and the mean number of generated
tokens per example. Unlike the likelihood-based multiple-choice evaluation
used in the main text, this protocol requires every setting to generate an
answer, making the four settings directly comparable in both performance and
generation cost.

\begin{table}[t]
\centering
\caption{\textbf{Four-setting comparison on Qwen3-8B.} We report accuracy and mean
generated tokens per example.}
\label{tab:cot_empirical}
\small
\begin{tabular}{@{}lcccccc@{}}
\toprule
& \multicolumn{2}{c}{2WikiMultiHopQA} & \multicolumn{2}{c}{MMLU} & \multicolumn{2}{c}{GSM8K} \\
\cmidrule(lr){2-3}\cmidrule(lr){4-5}\cmidrule(lr){6-7}
\textbf{Setting} & \textbf{Acc} & \textbf{Tokens} & \textbf{Acc} & \textbf{Tokens} & \textbf{Acc} & \textbf{Tokens} \\
\midrule
Base & 44.5 & 97 & 70.2 & 64 & 88.6 & 189 \\
\quad + CoT & 44.7 & 326 & 71.4 & 508 & 91.8 & 1002 \\
ReFlux (sync) & 49.5 & 120 & 71.2 & 129 & 90.4 & 263 \\
\rowcolor{E8F1FF}
\quad + CoT & 49.9 & 294 & 71.1 & 480 & 92.0 & 691 \\
\bottomrule
\end{tabular}
\end{table}

The four settings form a \(2\times2\) factorial design,
\(\{\text{CoT},\text{no-CoT}\}\times\{\text{ReFlux},\text{no-ReFlux}\}\),
allowing us to isolate the contributions of the two bottlenecks identified
by our analysis (Table \ref{tab:cot_empirical}). On 2WikiMultiHopQA, ReFlux improves accuracy from \(44.5\)
to \(49.5\), whereas CoT has almost no effect (\(44.5\rightarrow44.7\)).
On GSM8K, the pattern is reversed: CoT provides the larger gain
(\(88.6\rightarrow91.8\)), while ReFlux still improves performance to
\(90.4\). MMLU, which lies between these two regimes, shows smaller and more
balanced effects from the two interventions. These results are consistent
with the proposed bottleneck view: CoT primarily supplies additional serial
computation, whereas ReFlux primarily makes previously computed information
accessible across depth.

The computational footprints are also markedly asymmetric. CoT substantially
extends the generated reasoning trace, adding \(229\), \(444\), and \(813\)
tokens on 2WikiMultiHopQA, MMLU, and GSM8K, respectively. Thus, in this fully generative
setting, the two interventions differ not only in where they act but also in
how their additional computation is exposed at the output level: CoT
allocates computation through a longer sequence of generated intermediate
tokens, while ReFlux reuses internal computation without requiring additional
reasoning tokens.

More interestingly, adding ReFlux to CoT often \emph{shortens} the reasoning
trace rather than simply improving its accuracy. The mean generated length
drops from \(326\) to \(294\) tokens on 2WikiMultiHopQA, from \(508\) to
\(480\) on MMLU, and from \(1002\) to \(691\) on GSM8K, while accuracy is
maintained or improved in all three cases. This shortening is consistent
with a sub-additive interaction between the two interventions: once
previously computed information becomes accessible through ReFlux, part of
the serial reasoning that CoT would otherwise need to externalize can be
replaced by internal feedback. In other words, ReFlux does not merely add
another source of computation on top of CoT; it can change how much explicit
serial computation is needed downstream. The two mechanisms therefore remain
complementary, but their effects need not be additive: making existing
computation accessible can reduce the amount of additional computation that
must be generated explicitly.

\section{Experimental Details}
\label{app:details}

\subsection{Router Training Hyperparameters}
\label{app:router_hyperparameters}
\subsubsection{Architecture}
For routing block \(b\), we take the residual states after the final
Transformer block over the preceding token block, apply the backbone's
layer-normalization convention, and mean-pool over positions to obtain
\(q_b\in\mathbb{R}^{D}\). For the first routing block, where no preceding
block exists, we use a zero vector and therefore start from the router's
learned biases. In synchronous training and evaluation, \(q_b\) is computed
from the no-feedback reference traversal; in streaming evaluation, it is
computed from the feedback-augmented trajectory available at the end of the
preceding block. In parallel-prefill generation, the first decode block uses
the summary of the final prompt block. The router applies two fully connected layers of
width 512, each followed by GELU, with dropout \(0.1\) between the layers.
Two independent linear heads map the resulting representation to
\(\eta^{(b)},\rho^{(b)}\in\mathbb{R}^{|\mathcal{C}|}\), producing one gate
logit and one amplitude logit per candidate edge. Amplitudes use
\(\operatorname{softplus}\) and are clipped at \(0.25\) before the
architecture-specific write protocol is applied. The backbone parameters are
frozen; gradients update only the router.
For a backbone with \(L\)
Transformer blocks, the candidate set contains every deep-to-shallow pair of
block outputs,
\[
\mathcal C=\{(s,d):1\le d<s\le L\},
\qquad |\mathcal C|=\frac{L(L-1)}{2}.
\]
This gives 325, 561, and 1,128 candidate edges for Gemma3-1B, Gemma3-4B,
and Gemma3-12B, respectively, and 630 edges for both Qwen3-4B and Qwen3-8B.
The router therefore learns the useful depth intervals rather than inheriting
a geometry-based pruning rule.

\subsubsection{Optimization}
We train for 2,000 optimizer updates with AdamW
(\(\beta_1=0.9\), \(\beta_2=0.999\), \(\epsilon=10^{-8}\)), an initial
learning rate of \(10^{-4}\), and weight decay \(10^{-4}\). We use 100
linear warmup updates followed by cosine decay to \(10^{-5}\). 
Gradients are clipped to a global norm of 1.0. Backbone computation uses
bfloat16, while router parameters and optimizer states are maintained in
float32. Router training was performed with synchronous execution using data parallelism across eight NVIDIA RTX 5090 GPUs. For each backbone, the resulting router was reused unchanged for both synchronous and streaming inference. 

\subsubsection{Gating and regularization}
The Gumbel-Sigmoid temperature \(\tau\) is linearly annealed from 1.0 to 0.3
over the first 1,500 updates and held at 0.3 thereafter. At inference time,
an edge is active when its gate probability exceeds \(\xi=0.6\). In
Eq.~\ref{eq:router_loss}, both regularizers are averaged over routing blocks
and candidate edges; we set the sparsity coefficient to
\(\lambda_m=10^{-3}\) and the amplitude coefficient to
\(\lambda_\alpha=10^{-2}\).

\subsection{Efficiency Measurement}
\label{app:efficiency_protocol}

\subsubsection{Hardware and software}
We measure all efficiency numbers on a single NVIDIA RTX 5090 (32\,GB) with
exclusive GPU access. The software stack uses PyTorch 2.9.1+cu128, pinned
\texttt{transformers}, bf16 weights, and SDPA attention. Each run loads the
model directly onto the benchmarked device, and no other GPU workload is run
concurrently.

\subsubsection{Workload}
In our microbenchmark, a 1{,}024-token prompt is first
prefilled to initialize the KV state required by each method, after which the
model performs 128--192 single-token decode steps. Runtime is the end-to-end
wall-clock time of the decode phase, including both backbone computation and
feedback operations. We report latency as
milliseconds per decoded token, averaged over the final two of three
post-warm-up repetitions with explicit CUDA synchronization before and after
each repetition. For Synchronous and Original Recirculation, the measurement
includes advancing both the reference and feedback KV caches. Memory is measured by
\texttt{torch.cuda.max\_memory\_allocated} after warm-up, excluding the CUDA
context. Kernel counts and device time are attributed with
\texttt{torch.profiler} over 8 profiled decode steps.

\subsubsection{Analytic cost and memory}
We first separate the dominant backbone cost from the small auxiliary state
introduced by feedback. For a generated token, the base model costs \(2N\)
matmul FLOPs, where \(N\) is the number of model parameters. ReFlux-streaming
has the same backbone FLOPs as the base model: the feedback write is an
\(O(E\cdot D)\) elementwise operation and does not re-execute model weights.
ReFlux-synchronous and Original Recirculation perform two full traversals and
therefore cost \(2\times\); more generally, an iterated variant with \(r\)
feedback rounds costs \((r{+}1)\times\).

TF-Loop follows the out-of-the-box recipe recommended by its authors: a
contiguous window of \(W{=}4\) layers centered at fractional depth \(0.5\),
refined by 3-stage block-mode Runge--Kutta (\(K{=}3\), anchor
\(\beta{=}0.5\)). This placement coincides with the optimum of the original
window-position sweep and is shared across the two model families we evaluate.
Its analytic overhead is \(1+(K{-}1)W/L\), giving \(1.22\times\) at \(L{=}36\)
and \(1.31\times\) at \(L{=}26\).

Memory follows the same distinction. ReFlux changes the \emph{contents} of KV
entries but not the shape of an individual KV cache. Its persistent feedback
state consists of \(E\cdot D\cdot 2\) bytes of payload slots, plus transient
per-step captures. During autoregressive decoding, two-traversal methods keep
separate reference and feedback KV caches. Streaming advances only the
feedback trajectory and therefore needs a single KV cache, so its peak memory
stays essentially flat.

\subsubsection{Decode-time measurements}
Table~\ref{tab:efficiency} isolates the marginal cost of applying feedback
once autoregressive decoding is underway. The remaining gap between analytic
FLOPs and measured runtime is small and expected, because runtime also includes
fixed per-token costs that FLOPs does not track. ReFlux-streaming adds no
backbone matmuls and stays close to the base runtime (\(1.02\times\) on
Gemma3-1B and \(1.08\times\) on Qwen3-4B). The two-traversal methods,
Original Recirculation and ReFlux-synchronous, land near \(2\times\)
wall-clock cost. TF-Loop sits between them because it re-executes only a layer
window, measuring \(1.34\times\) on Gemma3-1B and \(1.21\times\) on Qwen3-4B.

The small residual overhead of ReFlux-streaming comes from our hook-based
prototype: the feedback write is launched from Python between existing kernel
calls. With the write fused into the layer loop or captured as a CUDA graph,
streaming approaches its analytic \(1.00\times\) cost. By contrast, the
overhead of TF-Loop and two-traversal feedback is intrinsic to re-executing
model weights. Implementation improvements should therefore mainly widen the
gap in favor of streaming, without changing the qualitative cost-quality
ordering reported in Table~\ref{tab:efficiency}.

\begin{figure*}[t]
\centering
\includegraphics[width=\textwidth]{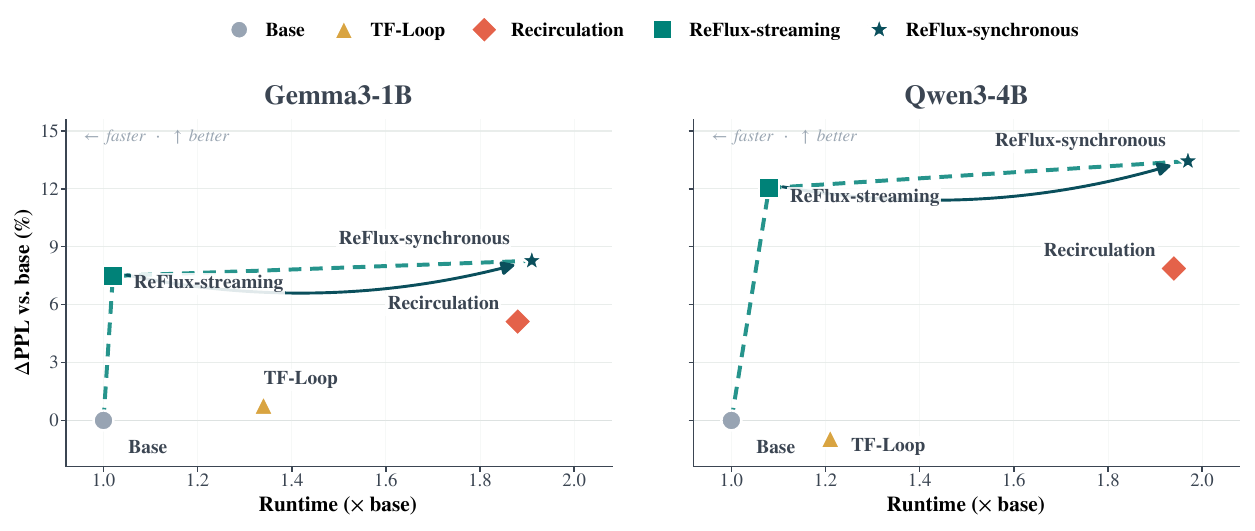}
\caption{Pareto view of generation efficiency and quality for Gemma3-1B and Qwen3-4B.
Streaming stays near the base runtime while retaining most of the PPL gain;
Synchronous reaches the highest-quality frontier at roughly $2\times$ cost;
Original Recirculation is dominated by the delta-based variants; and TF-Loop
remains lower in quality.}
\label{fig:eff_pareto}
\end{figure*}

Fig.\ref{fig:eff_pareto} visualizes the same tradeoff. Streaming stays
closest to the base-cost corner while preserving most of the quality gain;
Synchronous reaches the highest-quality frontier at roughly \(2\times\) cost;
Original Recirculation sits at a weaker point with similar cost; and TF-Loop,
although cheaper than the two-traversal schedules, does not dominate the
delta-based variants on either axis.

\subsubsection{Prefill and generation protocols}
\label{app:prefill_modes}
Autoregressive generation naturally separates into prompt prefill and
token-by-token decoding. For ReFlux-streaming, this separation creates two
useful operating points. \emph{Strict streaming} applies the cached increment
from position \(t-1\) before computing position \(t\), including throughout
the prompt. It preserves the recurrence in Eq.~\ref{eq:stream_payload} over
the entire sequence, but necessarily serializes prompt prefill over tokens.
\emph{Parallel-prefill streaming} instead performs the prompt with the
standard batched causal prefill, initializes the payload cache from the final
prompt position, and activates the same chained update only during
autoregressive decoding. It therefore retains token-to-token feedback during
answer generation without introducing a serial prefill cost.

The two operating points answer different questions. Strict full-sequence
streaming measures the quality effect of enforcing the recurrence over every
scored token, including prompt tokens. Parallel-prefill streaming is the
generation serving protocol: it preserves standard one-pass prompt processing
and applies ReFlux on the already sequential decoding path. All ReFlux routers
are trained under synchronous execution and then reused unchanged in the
streaming settings, so differences between these columns reflect execution
schedule rather than router retraining.

The main tables follow this distinction. Table~\ref{tab:main-ppl} evaluates
language modeling under \emph{strict full-sequence streaming}, so its streaming
PPL measures the recurrence itself rather than making a latency claim about
parallel prefill. Table~\ref{tab:main-accuracy} uses
\emph{parallel-prefill streaming} for open-ended generation benchmarks
(GSM8K, HotpotQA, and 2WikiMultiHopQA): the prompt is encoded once using the
standard batched prefill path, and feedback is activated during
autoregressive decoding. The multiple-choice benchmarks (MMLU, ARC-Easy,
ARC-Challenge, HellaSwag, and Winogrande) contain no autoregressive decoding
phase; candidates are selected by length-normalized conditional likelihood, so
feedback can only act during the strict sequential scoring pass.

\subsubsection{End-to-end prefill/decode decomposition}
Table~\ref{tab:prefill_decode_e2e} complements the decode-only measurement in
Table~\ref{tab:efficiency}. Instead of isolating per-token decode overhead, it
decomposes a complete prompt-to-answer request into prefill, time-to-first-token
(TTFT), and decoding. This separates the cost of enforcing recurrence during
prompt processing from the cost of using the same recurrence during
autoregressive decoding.

\begin{table}[t]
\centering
\caption{Prefill/decode decomposition on a supplementary multi-hop generation
workload. This table uses Gemma3-4B on 128 HotpotQA and 2WikiMultiHopQA
prompts (mean length 1107 tokens), batch size 1, and 24 teacher-forced decode
steps. Ratios are relative to Base within this workload.}
\label{tab:prefill_decode_e2e}
\scriptsize
\setlength{\tabcolsep}{2.0pt}
\renewcommand{\arraystretch}{1.06}
\resizebox{\columnwidth}{!}{%
\begin{tabular}{@{}lllrrrrrrr@{}}
\toprule
\textbf{Schedule} & \textbf{Prefill} & \textbf{Decode} &
\textbf{Prefill (ms)} & \textbf{TTFT (ms)} &
\textbf{Decode (ms/tok)} & \textbf{Total (s)} &
\textbf{Prefill \(\times\)} & \textbf{Decode \(\times\)} &
\textbf{E2E \(\times\)} \\
\midrule
Base & Parallel, 1 pass & Standard
& 69.4 & 104.9 & 32.67 & 0.85 & 1.00 & 1.00 & 1.00 \\
Strict Stream & Sequential & Chained
& 35378.8 & 35412.1 & 33.03 & 36.17 & 510.08 & 1.01 & 42.38 \\
Parallel-Prefill Stream & Parallel, 1 pass & Chained
& 70.1 & 105.9 & 33.76 & 0.87 & 1.01 & 1.03 & 1.02 \\
Synchronous & Parallel, 2 pass & 2 pass
& 134.0 & 201.0 & 66.24 & 1.72 & 1.93 & 2.03 & 2.02 \\
Orig. Recirculation & Parallel, 2 pass & 2 pass
& 134.7 & 204.8 & 67.36 & 1.75 & 1.94 & 2.06 & 2.05 \\
TF-Loop & Parallel, looped window & Looped
& 82.5 & 124.1 & 39.26 & 1.02 & 1.19 & 1.20 & 1.20 \\
\bottomrule
\end{tabular}}
\end{table}

The decomposition is straightforward. Strict streaming has almost no
per-token decode overhead (\(1.01\times\)), but enforcing recurrence
throughout the prompt turns prefill into a token-by-token computation. This cost is a
property of the strict execution schedule, not an intrinsic decode overhead of
ReFlux. Parallel-prefill streaming avoids the serialization: its prompt
prefill and TTFT remain essentially at the base-model level while the same
chained update is preserved during decoding. Consequently, the streaming
generation results in Table~\ref{tab:main-accuracy} do not depend on serial
prompt processing. Synchronous execution performs two passes and therefore
incurs roughly \(2\times\) prefill and decode cost; Original Recirculation has
similar accounting. TF-Loop falls between these regimes at approximately
\(1.2\times\) end-to-end cost.

\subsubsection{Recurrence Without Repeated Traversal}
\label{sec:recurrence_without_repeated_traversal}
Taken together, these measurements show that recurrence does not have to mean another backbone traversal. Synchronous ReFlux realizes recurrence by re-executing the model, whereas ReFlux-streaming carries the feedback increment forward and incorporates it into the next token's existing traversal. Streaming therefore adds recurrent computation along the autoregressive trajectory without adding another full pass through the weights.

These observations make parallel-prefill streaming particularly well suited to practical generation serving. It leaves the inherently parallel prompt computation unchanged and places the recurrent update entirely on the token-by-token decoding path, where sequential computation already exists. In this sense, ReFlux does not introduce additional serialization to realize recurrence, it aligns the recurrent computation with the serialization that is already intrinsic to autoregressive decoding. The resulting execution schedule preserves standard batched prefill while augmenting each subsequent decode step with a lightweight feedback update, making parallel-prefill streaming the natural deployment operating point for ReFlux. In general, this suggests a useful systems principle for recurrent augmentation of Transformers: when recurrence can be carried along the existing autoregressive trajectory, it can provide additional computation without converting an otherwise parallel phase of inference into a sequential one.

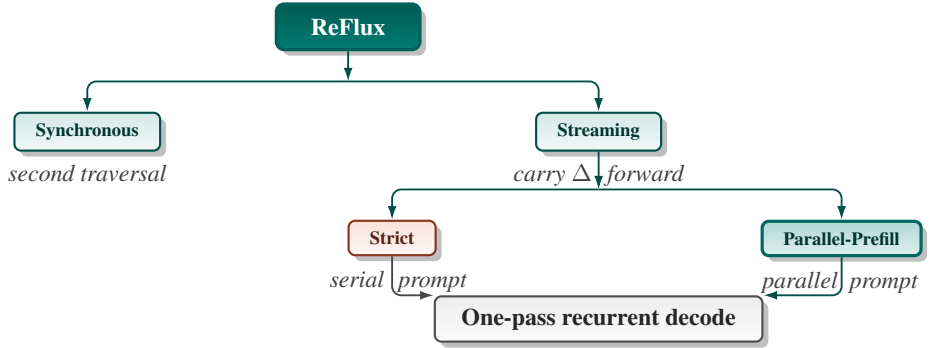
\begin{figure}[t]
    \centering
    \begin{tikzpicture}[
        every node/.style={
            align=center,
            font=\small
        },
        root/.style={
            draw=BlueGreen!75!black, very thick,
            top color=BlueGreen!70!black, bottom color=BlueGreen!95!black,
            text=white, rounded corners=3pt,
            inner xsep=13pt, inner ysep=5.5pt,
            font=\small\bfseries, drop shadow
        },
        branch/.style={
            draw=#1!60!black, thick,
            top color=#1!18, bottom color=#1!5,
            text=#1!40!black, rounded corners=2.5pt,
            inner xsep=8pt, inner ysep=4pt,
            font=\scriptsize\bfseries, drop shadow
        },
        detail/.style={
            font=\footnotesize\itshape,
            text=gray!50!black
        },
        accent/.style={
            branch=BlueGreen,
            draw=BlueGreen!80!black, very thick,
            top color=BlueGreen!32, bottom color=BlueGreen!12
        },
        shared/.style={
            draw=gray!60!black, thick,
            top color=gray!14, bottom color=gray!4,
            text=gray!30!black, rounded corners=2.5pt,
            inner xsep=10pt, inner ysep=4.5pt,
            font=\small\bfseries, drop shadow
        },
        arrow/.style={
            -{Latex[length=1.8mm,width=1.2mm]},
            line width=0.7pt,
            draw=BlueGreen!60!black,
            rounded corners=1.2mm
        },
        arrowgray/.style={
            -{Latex[length=1.8mm,width=1.2mm]},
            line width=0.7pt,
            draw=gray!55!black,
            rounded corners=1.2mm
        }
    ]

    \node[root] (root) {ReFlux};

    \node[branch=BlueGreen, below left=8mm and 15mm of root] (sync)
        {Synchronous};
    \node[branch=BlueGreen, below right=8mm and 15mm of root] (stream)
        {Streaming};

    \coordinate (rootfork) at ([yshift=-4mm]root.south);

    \draw[arrow] (root.south) -- (rootfork);
    \draw[arrow] (rootfork) -| (sync.north);
    \draw[arrow] (rootfork) -| (stream.north);

    \node[detail, below=0.5mm of sync]
        {second traversal};

    \node[detail, below=0.5mm of stream]
        {carry $\Delta$\hspace{0.4em} forward};

    \node[branch=RedOrange, below left=9mm and 13mm of stream] (strict)
        {Strict};
    \node[accent, below right=9mm and 13mm of stream] (parallel)
        {Parallel-Prefill};

    \coordinate (streamfork) at ([yshift=-5mm]stream.south);

    \draw[arrow] (stream.south) -- (streamfork);
    \draw[arrow] (streamfork) -| (strict.north);
    \draw[arrow] (streamfork) -| (parallel.north);

    \node[detail, below=0.6mm of strict, xshift=0.8mm]
        {serial\hspace{0.6em}prompt};
    \node[detail, below=0.6mm of parallel, xshift=-0.2mm]
        {parallel\hspace{0.5em}prompt};

    \node[shared, below=19mm of stream] (decode)
        {One-pass recurrent decode};

    \draw[arrowgray] (strict.south) |- (decode.north west);
    \draw[arrow] (parallel.south) |- (decode.north east);

    \end{tikzpicture}

    \caption{
    ReFlux execution paths. Synchronous recurrence requires a second backbone
    traversal, whereas streaming carries the feedback increment across tokens.
    The two streaming variants differ only in prompt processing and share the
    same one-pass recurrent decoding mechanism.
    }
    \label{fig:reflux_streaming_modes}
\end{figure}

\subsection{Routing-Block Sensitivity}
\label{app:block_sensitivity}
The block length \(w\) determines how long one routing decision is reused.
We retrain the Gemma3-4B router for
\(w\in\{32,64,128,256,512\}\), changing no other setting. Short blocks can
respond more quickly to changes in context, but call the router more often;
long blocks are cheaper, but force more tokens to share the same route.
Table~\ref{tab:block-sensitivity} reports this trade-off under streaming
execution. For a 1,024-token window, the five settings make 32, 16, 8, 4, and
2 routing decisions, respectively.

\begin{table}[t]
\centering
\caption{Sensitivity to routing-block length on Gemma3-4B using
ReFlux-streaming. Runtime is normalized to the base model; lower C4 PPL and
higher ARC-C accuracy are better.}
\label{tab:block-sensitivity}
\small
\setlength{\tabcolsep}{4.2pt}
\renewcommand{\arraystretch}{1.05}
\begin{tabular}{@{}ccccc@{}}
\toprule
\textbf{Block \(w\)} & \textbf{C4 PPL \(\downarrow\)} &
\textbf{ARC-C \(\uparrow\)} & \textbf{\(\bar\kappa\)} &
\textbf{Runtime (\(\times\))} \\
\midrule
32  & 21.1 & 58.1 & 2.73 & 1.08 \\
64  & \textbf{21.0} & \textbf{58.2} & 2.66 & 1.06 \\
\textbf{128} & 21.2 & 58.0 & 2.58 & 1.04 \\
256 & 21.4 & 57.8 & 2.31 & 1.04 \\
512 & 21.8 & 57.4 & 2.44 & 1.03 \\
\bottomrule
\end{tabular}
\end{table}
Quality changes little between \(w=32\) and \(128\), but begins to fall once
one route is shared over 256 or more tokens. We therefore use \(w=128\) during language modeling: it
stays on the quality plateau while invoking the router \(4\times\) less often
than \(w=32\). For benchmark evaluation, we retain \(w=128\) by default, reducing it to \(32\) or \(4\) when necessary to ensure that even the shortest inputs receive at least one context-conditioned routing step.

\subsection{Routing-Threshold Sensitivity}
\label{app:threshold_sensitivity}
The threshold \(\xi\) turns the router's soft gate probabilities into an
active graph. We sweep \(\xi\in\{0.4,0.5,0.6,0.7,0.8\}\) at inference time
using the same trained Gemma3-4B router. Lower thresholds retain more
candidate edges; higher thresholds make the route sparser. No retraining or
other hyperparameter change is made in this comparison.

\begin{table}[t]
\centering
\caption{Sensitivity to the inference-time routing threshold on Gemma3-4B
using ReFlux-streaming. The router is not retrained between settings. Runtime
is normalized to the base model.}
\label{tab:threshold-sensitivity}
\small
\setlength{\tabcolsep}{4.2pt}
\renewcommand{\arraystretch}{1.05}
\begin{tabular}{@{}ccccc@{}}
\toprule
\textbf{Threshold \(\xi\)} & \textbf{C4 PPL \(\downarrow\)} &
\textbf{ARC-C \(\uparrow\)} & \textbf{\(\bar\kappa\)} &
\textbf{Runtime (\(\times\))} \\
\midrule
0.4 & 21.5 & 57.6 & 4.61 & 1.06 \\
0.5 & \textbf{21.1} & \textbf{58.1} & 3.37 & 1.06 \\
\textbf{0.6} & 21.2 & 58.0 & 2.58 & 1.04 \\
0.7 & 21.5 & 57.7 & 1.74 & 1.03 \\
0.8 & 22.0 & 57.1 & 0.91 & 1.03 \\
\bottomrule
\end{tabular}
\end{table}
The best region is broad: moving from \(\xi=0.5\) to \(0.6\) removes roughly
one active edge on average with almost no loss in quality. Below this range,
weakly scored writes begin to interfere; above it, useful increments are
discarded. We choose \(\xi=0.6\) as the sparser point in this stable region.

\subsection{Routing Details}
\label{app:routing_details}
\subsubsection{Routing statistics}
We conduct all routing-behavior analyses on Gemma3-1B. Fig.\ref{fig:routing-behavior}
visualizes the block-level pattern, and Table~\ref{tab:routing-summary}
summarizes the same measurements by task family.
We define block difficulty as the mean token NLL $\ell_b$
under the unmodified backbone. Because $q_{b+1}$ summarizes block $b$, we pair
$\ell_b$ with the subsequent active-edge count $\kappa_{b+1}$ and omit the
first block. Panel (b) reports the activation frequency
$f_e=B^{-1}\sum_{b=1}^{B}\mathbbm{1}\{e\in A_b\}$ for each candidate edge.
Let $\alpha^{(b)}_e=m_e^{(b)}a_e^{(b)}$ be the realized write strength of edge
$e$ in block $b$. At inference time, the active set is
$A_b=\{e:m_e^{(b)}=1\}=\{e:\sigma(\eta_e^{(b)})>\xi\}$, with $\xi=0.6$ in
all reported runs. Then
$\kappa_b=|A_b|$ counts active edges. To measure route concentration before
hard thresholding, we define the soft write mass
$w_e^{(b)}=\sigma(\eta_e^{(b)})a_e^{(b)}$ and its normalization
$\pi_e^{(b)}=w_e^{(b)}/\sum_{e'\in\mathcal{C}}w_{e'}^{(b)}$. The route entropy is
$H_b=-\sum_{e\in\mathcal{C}}\pi_e^{(b)}\log\pi_e^{(b)}$. We normalize edge
span by model depth and define the mean depth span as
$\delta_b=\frac{1}{|A_b|}\sum_{e\in A_b}(s_e-d_e)/L$, where $(s_e,d_e)$ are
the source and target layers of edge $e$. For route stability, we set the overlap
rank to $k=3$ and compare the top-$k$ soft-route sets by Jaccard overlap:
$|S_b^{(k)}\cap S_{b'}^{(k)}|/|S_b^{(k)}\cup S_{b'}^{(k)}|$, where
$S_b^{(k)}$ contains the $k$ candidate edges with the largest soft write mass
$w_e^{(b)}$ in block $b$, before hard thresholding.
Same-family overlap averages this score over block pairs from the same task
family, while cross-family overlap averages it over pairs from different
families.

Table~\ref{tab:routing-summary} shows that the router remains sparse but changes its support with
the input. Multi-hop tasks activate the most edges and have the largest depth
span, consistent with the need to propagate intermediate evidence across
longer computation paths. In contrast, HellaSwag and Winogrande use fewer and
shorter routes, matching their more local completion-style structure. Across
all families, same-family overlap is substantially higher than cross-family
overlap, indicating that related inputs reuse similar routing motifs while
different task families induce different feedback patterns.

\begin{table}[t]
\centering
\caption{Routing summary by task family on Gemma3-1B. \(\kappa\) is the
active-edge count, \(H\) is route entropy, overlap compares top-$k$ soft-route sets,
and depth span is the source-minus-target layer distance normalized by model
depth.}
\label{tab:routing-summary}
\small
\setlength{\tabcolsep}{3.4pt}
\renewcommand{\arraystretch}{1.04}
\begin{tabular}{@{}lccccc@{}}
\toprule
Task family & $\kappa$ & $H$ & Same ovlp. & Cross ovlp. & Depth span \\
\midrule
C4 / WebTextLike & 2.18 & 1.09 & 0.64 & 0.19 & 0.27 \\
BookSum / PG19 & 2.87 & 1.23 & 0.58 & 0.33 & 0.36 \\
MMLU / ARC & 2.33 & 1.14 & 0.67 & 0.30 & 0.24 \\
HellaSwag / Winogrande & 1.96 & 1.06 & 0.69 & 0.38 & 0.22 \\
GSM8K & 2.54 & 1.19 & 0.60 & 0.34 & 0.31 \\
HotpotQA / 2WikiMultiHopQA & 3.12 & 1.34 & 0.54 & 0.22 & 0.42 \\
\midrule
Mean & 2.50 & 1.18 & 0.62 & 0.29 & 0.30 \\
\bottomrule
\end{tabular}
\end{table}

\label{app:evaluation_protocol}

\subsection{evaluation protocol}
We evaluate reasoning performance using widely adopted evaluation protocols implemented in the \texttt{lm-evaluation-harness}. We keep the prompt construction, few-shot demonstrations, and decoding settings fixed across the base and ReFlux variants, so that performance differences can be attributed to the proposed inference-time intervention. All models are evaluated as pretrained base language models without instruction tuning or chat templates. For the multiple-choice benchmarks, we use 5-shot evaluation on MMLU, 0-shot evaluation on ARC-Easy, 25-shot evaluation on ARC-Challenge, 10-shot evaluation on HellaSwag, and 5-shot evaluation on WinoGrande. For these closed-set tasks, 
the prediction is the candidate with the highest length-normalized
conditional log-likelihood. 

For GSM8K, we follow the 5-shot generation-based evaluation protocol provided by \texttt{lm-evaluation-harness}. The model generates a solution autoregressively with greedy decoding (\texttt{temperature}=0), and the final numerical answer is extracted from the generated response for exact-match evaluation. For HotpotQA and 2WikiMultiHopQA, we use the context-grounded open-ended question-answering format described in Appendix~\ref{app:prompt_templates}, without task-specific fine-tuning or external retrieval, and report exact match (EM). Unless otherwise specified, all benchmarks are evaluated on their original test or evaluation splits using the same prompt construction and decoding configuration for all model variants.

For comparisons with the original Recirculation method, we follow the evaluation protocol reported in its publicly available initial release, including the corresponding preprocessing and loss-computation procedures. The TF-Loop baseline is implemented from the recipe recommended by its authors: a contiguous window of $W{=}4$ layers centered at mid-depth is re-executed with three-stage block-mode Runge--Kutta refinement ($K{=}3$, anchor $\beta{=}0.5$). Concretely, each forward pass captures the activation entering the window, re-runs the window layers $K$ times while collecting successive exit increments, combines them with the Runge--Kutta weights, and injects the combined update at the window exit; the layers outside the window, attention, and the key/value cache are left untouched, and no weights are updated.

\section{Ablations}
\subsection{Ablation Protocol}
\label{app:ablation_protocol}

We conduct controlled ablations on Gemma3-4B to isolate the contributions of the feedback payload, edge-wise strength allocation, and input-conditioned routing. Unless otherwise stated, all settings use the same frozen backbone, candidate edge set, training data, optimization procedure, routing frequency, normalization operator, and inference budget as ReFlux. The ablations are trained with the same router parameterization and number of optimization steps.

\paragraph{Reference configuration.}
In the standard ReFlux configuration, each candidate edge $e=(s_e,d_e)$ carries the depth increment

$$
\Delta_e(t)
=
z^{(0)}_{s_e}(t)-z^{(0)}_{d_e}(t),
$$

where $z^{(0)}$ denotes the backbone representations before feedback. The router produces a binary gate $m_e^{(b)}$ and a non-negative amplitude $a_e^{(b)}$ for each edge at routing block $b$, giving

$$
\alpha_e^{(b)}
=
m_e^{(b)}a_e^{(b)}.
$$

Let \(h_d(t)\) denote the current residual stream at target depth \(d\)
immediately before the write. The resulting update is

$$
\bar h_d(t)
=
h_d(t)
+
\sum_{e\in \mathcal I_d^{(b)}}
\alpha_e^{(b)}
N_{h_d(t)}
\!\left(\Delta_e(t)\right),
$$

where \(N\) denotes the same write-back normalization used by ReFlux. Thus,
the payload is formed from the reference traversal, while the write is
normalized against the current target stream. ReFlux jointly learns
\emph{what} to propagate through the incremental payload, \emph{which} edges
to activate, and \emph{how strongly} to apply each active edge.

\paragraph{Full-state payload.}
\textbf{Learned-full} replaces the incremental payload with the complete source representation while leaving the routing mechanism unchanged. Specifically,

$$
p_e(t)=z^{(0)}_{s_e}(t),
$$

instead of

$$
p_e(t)=\Delta_e(t)
=z^{(0)}_{s_e}(t)-z^{(0)}_{d_e}(t).
$$

The router remains input-conditioned and continues to learn both edge gates and edge-wise amplitudes. The resulting update is therefore

$$
\bar h_d(t)
=
h_d(t)
+
\sum_{e\in \mathcal I_d^{(b)}}
\alpha_e^{(b)}
N_{h_d(t)}
\!\left(z^{(0)}_{s_e}(t)\right).
$$

This setting tests whether the benefit of ReFlux arises specifically from propagating newly accumulated computation, rather than from feeding back the entire source state.

\paragraph{Uniform feedback strength.}
\textbf{Uniform-strength} retains the incremental payload and the input-conditioned edge support, but removes edge-wise strength adaptation. Let

$$
A_b=\{e:m_e^{(b)}=1\}
$$

be the active edge set at routing block $b$. Instead of using the learned amplitudes $a_e^{(b)}$, all active edges are assigned a common strength,

$$
\alpha_e^{(b)}=\bar\alpha_b,
\qquad e\in A_b,
$$

where

$$
\bar\alpha_b
=
\frac{1}{|A_b|}
\sum_{e\in A_b}
\alpha_e^{(b)}
$$

matches the mean active-edge strength of the corresponding learned router. The support $A_b$ is still selected by the input-conditioned gate head, and the payload remains

$$
p_e(t)=\Delta_e(t).
$$

Consequently, this ablation isolates the contribution of learned edge-wise strength allocation.

\paragraph{Fixed incremental route.}
\textbf{Fixed-Delta} retains both the incremental payload and learned edge-wise strengths, but removes input-conditioned changes in the edge support. We first identify a fixed edge configuration

$$
E_{\mathrm{fixed}}\subseteq C
$$

from the best-performing edge configuration in the depth-geometry sweep described in Sec.~\ref{sec:scaling}. The selected edge set is then used for every input and every routing block:

$$
m_e^{(b)}
=
\begin{cases}
1,&e\in E_{\mathrm{fixed}},\\
0,&e\notin E_{\mathrm{fixed}}.
\end{cases}
$$

For selected edges, the amplitude remains input-conditioned and is produced by the learned amplitude head. Hence,

$$
p_e(t)=\Delta_e(t),
\qquad
\alpha_e^{(b)}=a_e^{(b)}
\quad\text{for }e\in E_{\mathrm{fixed}}.
$$

The size of $E_{\mathrm{fixed}}$ is matched to the active-edge budget of ReFlux. This design isolates the value of dynamically adapting the edge support while retaining the same incremental payload and strength adaptation.

\paragraph{Geometry-based heuristic route.}
\textbf{Heuristic-Delta} retains the incremental payload and learned edge-wise strengths, but replaces learned input-conditioned support selection with a fixed representation-geometry heuristic. The heuristic uses only backbone representations from a calibration set and does not use router scores, downstream task labels, or evaluation data.

For each candidate edge $e=(s_e,d_e)$, we compute its average cross-depth cosine similarity on the C4 calibration set $\mathcal D_{\mathrm{cal}}$:

$$
C_e
=
\mathbb{E}_{x\sim\mathcal D_{\mathrm{cal}}}
\left[
\cos
\left(
z^{(0)}_{s_e}(x),
z^{(0)}_{d_e}(x)
\right)
\right].
$$

We then define the representation deviation score

$$
D_e=1-C_e.
$$

Edges with larger $D_e$ correspond to source--target pairs whose representations exhibit larger angular deviation across depth. We select the $K$ candidate edges with the largest deviation scores,

$$
E_{\mathrm{heuristic}}
=
\operatorname{TopK}_{e\in C}D_e,
$$

where $K$ is chosen to match the active-edge budget of ReFlux. The selected support is fixed across all inputs and routing blocks. For the selected edges, the payload remains

$$
p_e(t)=\Delta_e(t),
$$

while their amplitudes are still produced by the learned amplitude head. The heuristic is therefore a static alternative to learned route selection that directly exploits the cross-depth representation geometry identified previously.

\subsection{Budget-Matched Adaptation Baselines}
\label{app:budget_matched}

To separate the benefit of increment routing from the generic benefit of
fitting a small module, we compare several adaptation mechanisms on Qwen3-4B.
They see the same calibration windows and use the same optimizer, update
count, and language-modeling objective as ReFlux; the pretrained backbone
remains frozen.

\begin{enumerate}[leftmargin=1.5em,itemsep=1pt,topsep=2pt]
\item \textbf{Learned constant mixing} learns one gate and one amplitude for
each candidate edge, but uses the resulting route for every input
(\(2|\mathcal C|=1{,}260\) parameters on Qwen3-4B).
\item \textbf{Conditional-vector mixing} follows the adaptive variant of
\citet{mozer2026recirculation}, using an MLP over \([z_s;z_d]\) to produce
per-dimension mixing coefficients.
\item \textbf{LoRA-r8 \citep{hu2021lora}}: rank-8 LoRA adapters  on the attention $q,v$ projections,
trained with the same calibration loss.

\end{enumerate}

\begin{table}[t]
\centering
\caption{Budget-matched adaptation baselines on Qwen3-4B. All methods use
the same calibration data, optimizer, and update count; the backbone is
frozen throughout. ReFlux is evaluated with synchronous execution.}
\label{tab:budget_matched}
\small
\setlength{\tabcolsep}{5pt}
\begin{tabular}{@{}lcccc@{}}
\toprule
 & Params & Train time & C4 PPL $\downarrow$ & ARC-C $\uparrow$ \\
\midrule
Base                         & 0     & --    & 21.6 & 53.8 \\
Learned constant             & 1,260 & 1.3 h & 20.9 & 54.2 \\
Conditional-vector MLP       & 5.2M  & 1.8 h & 19.9 & 54.9 \\
LoRA-r8 ($q,v$)              & 2.9M  & 2.2 h & 19.6 & 55.1 \\
\rowcolor{E8F1FF}
ReFlux router                & 2.2M  & 1.5 h & \textbf{18.7} & \textbf{55.8} \\
\bottomrule
\end{tabular}
\end{table}

Learned constants recover only a small part of the improvement, reducing PPL
by 0.7 and raising ARC-C by 0.4 points. Conditioning the intervention or
adapting attention with LoRA is stronger, but both remain behind ReFlux.
In particular, ReFlux improves over LoRA by another 0.9 PPL and 0.7 accuracy
points despite using fewer parameters and less training time. The comparison
therefore attributes the gain not simply to adding trainable parameters, but
to choosing which endogenous depth increments should be reused for the
current context.

\section{Complete Multi-hop Case}
\label{app:complete_case}

\casebox{TrophyColor}{Question and task structure}{%
\textbf{Question:} \emph{Where was the director of film \textit{Rough Going}
born?}\\[-1pt]
\textbf{Gold answer:} \textcolor{TrophyColor}{New Hyde Park, New York}\\[-1pt]
\textbf{Required chain:} \textit{Rough Going}
$\xrightarrow{\text{directed by}}$ \textit{Wally Van}
$\xrightarrow{\text{born in}}$ \textcolor{TrophyColor}{New Hyde Park, New York}.
}

We give the complete input and recorded outputs for the multi-hop
example used in Sec.~\ref{sec:understanding-reflux}. The case is from the
2WikiMultiHopQA validation split (example 7688). Relevant passages
are interleaved with distractors that share the surface words ``director'' and
``born''.

\casebox{BlueGreen}{Complete context}{%
\small
\begin{itemize}[leftmargin=1.35em,itemsep=2pt,topsep=2pt]
\item[\textbf{p0}] \textbf{Olav Aaraas:} Norwegian historian and museum
director, born in Fredrikstad. \emph{Distractor.}
\item[\textbf{p1}] \textbf{Peter Levin:} American director of film and
television. \emph{Distractor.}
\item[\textbf{p2}] \textbf{Wally Van:} ``Wally Van (1880--1974) was an
American actor and film director. He was born in \textcolor{TrophyColor}{New
Hyde Park, New York} and died in Englewood, New Jersey.'' \emph{Evidence for
hop 2.}
\item[\textbf{p3}] \textbf{Pearl Going:} New Zealand socialite.
\emph{Distractor sharing ``Going''.}
\item[\textbf{p4}] \textbf{Jesse E. Hobson:} director of SRI International.
\emph{Distractor.}
\item[\textbf{p5}] \textbf{Rough Going:} ``\textit{Rough Going} is a 1925
silent film western \textcolor{SuitcaseColor}{directed by Wally Van} and
starring Franklyn Farnum.'' \emph{Evidence for hop 1.}
\item[\textbf{p6}] \textbf{Joanna Going:} American actress. \emph{Distractor
sharing ``Going''.}
\item[\textbf{p7}] \textbf{Ian Barry:} Australian director of film and
television. \emph{Distractor.}
\item[\textbf{p8}] \textbf{S.N. Mathur:} Director of the Indian Intelligence
Bureau. \emph{Distractor.}
\item[\textbf{p9}] \textbf{Brian Kennedy:} art museum director and director of
the Toledo Museum of Art. \emph{Distractor.}
\end{itemize}
}

The two evidence passages are separated in the context: p5 identifies the
director of the queried film, whereas p2 supplies that person's birthplace.
The bridge entity \textit{Wally Van} must therefore be retained while the
model moves from the first passage to the second hop.

\casebox{SuitcaseColor}{Recorded model outputs}{%
\begin{tcolorbox}[enhanced, colback=red!4, colframe=red!45!black,
  boxrule=0.5pt, arc=2pt, left=5pt, right=5pt, top=4pt, bottom=4pt,
  fontupper=\small\ttfamily, fonttitle=\small\bfseries\sffamily,
  coltitle=white, colbacktitle=red!45!black,
  title={Base model --- incorrect}]
The question is asking where the director of the film Rough Going
was born. From the context, Rough Going was directed by Wally Van.
Therefore, the relevant person is Wally Van. Let me check again, the director of Rough Going is indeed Wally Van.

Wally Van was an American actor and film director.

\textbf{Wally Van.}
\end{tcolorbox}

\vspace{3pt}

\begin{tcolorbox}[enhanced, colback=green!5, colframe=green!35!black,
  boxrule=0.5pt, arc=2pt, left=5pt, right=5pt, top=4pt, bottom=4pt,
  fontupper=\small\ttfamily, fonttitle=\small\bfseries\sffamily,
  coltitle=white, colbacktitle=green!35!black,
  title={ReFlux --- correct}]
The film Rough Going was directed by Wally Van. I therefore need to
follow the relation from Rough Going to its director, Wally Van, and
then determine where Wally Van was born. The passage about Wally Van
states that he was born in New Hyde Park, New York.

\textbf{New Hyde Park, New York.}
\end{tcolorbox}

}

The baseline reaches the bridge entity but does not complete the second hop in
its recorded continuation. ReFlux produces the complete relation in its reasoning trajectory.

This behavior is consistent with prior analyses of recurrent-depth transformers, which show that intermediate entities may only become recoverable at deep layers, too late to support subsequent reasoning, and that recurrence can repair this limitation by re-executing the network \citep{kohli2026loop}. ReFlux instead makes the intermediate result available at inference time by propagating the newly computed increment back to an earlier, queryable state. More broadly, this mechanism is consistent with our aggregate results: ReFlux yields its strongest gains on multi-hop tasks, where later computation must access conclusions formed at earlier positions.

\section{ReFlux Execution Algorithm}
\label{app:reflux_algorithm}
\begin{algorithm}[t]
\caption{Execution workflow of ReFlux}
\label{alg:reflux}
\Input{Frozen Transformer $f_\theta$, router $g_\phi$, candidate edges
$\mathcal{C}$, schedule $s$.}
\Output{Logits $\{\ell_t\}$ and payload cache $\{P_e\}_{e\in\mathcal{C}}$.}

\For{\rm{input sequence} $x$}{

\tcc{\textcolor{RefColor}{Shared initialization}}
$P_e\leftarrow 0\;(\forall e\in\mathcal{C})$, \quad
$\mathrm{KV}\leftarrow\emptyset$\;

\uIf{$s=\mathrm{parallel\mbox{-}prefill\ stream}$}{

\tcc{\textcolor{RefColor}{Parallel prompt prefill}}
$(\mathrm{KV},z^{(0)}(T))\leftarrow f_\theta(x_{\le T};p_e=0)$
\tcp*[r]{standard batched prefill}
$P_e\leftarrow z_{s_e}^{(0)}(T)-z_{d_e}^{(0)}(T)\;(\forall e\in\mathcal{C})$\;
$q_1\leftarrow\operatorname{Summary}(x_{\le T})$\;

\tcc{\textcolor{RefColor}{Streaming decode}}
\For{\rm{decode block} $b$}{
$\alpha^{(b)}\leftarrow\operatorname{Route}_\phi(q_b)$
\tcp*[r]{Eqs.~\ref{eq:routing_logits}--\ref{eq:threshold_route}}
\For{\rm{generated token} $t\in\mathcal{T}_b$}{
$(\hat z(t),\ell_t,\mathrm{KV})\leftarrow
f_\theta(x_t;\mathrm{KV},P,\alpha^{(b)})$
\tcp*[r]{Eq.~\ref{eq:delta_write}}
$P_e\leftarrow\hat z_{s_e}(t)-\hat z_{d_e}(t)\;(\forall e\in\mathcal{C})$
\tcp*[r]{Eq.~\ref{eq:stream_payload}}
}
$q_{b+1}\leftarrow\operatorname{Summary}(\mathcal{T}_b)$\;
}
}
\Else{

\tcc{\textcolor{RefColor}{Full-sequence scoring}}
\For{\rm{routing block} $b$}{
$q_b\leftarrow
\begin{cases}
0, & b=1,\\
\operatorname{Summary}(\mathcal{T}_{b-1}), & b>1,
\end{cases}$
\quad
$\alpha^{(b)}\leftarrow\operatorname{Route}_\phi(q_b)$\;

\uIf{$s=\mathrm{synchronous}$}{
\tcc{\textcolor{RefColor}{Synchronous two-pass traversal}}
$z^{(0)}(\mathcal{T}_b)\leftarrow f_\theta(x_{\mathcal{T}_b};p_e=0)$\;
$p_e(t)\leftarrow z_{s_e}^{(0)}(t)-z_{d_e}^{(0)}(t)$
\quad $(t\in\mathcal{T}_b,\ e\in\mathcal{C})$\;
$(\hat z(\mathcal{T}_b),\ell_{\mathcal{T}_b})\leftarrow
f_\theta(x_{\mathcal{T}_b};p(t),\alpha^{(b)})$
\tcp*[r]{Eq.~\ref{eq:delta_write}}
}
\ElseIf{$s=\mathrm{strict\ stream}$}{
\tcc{\textcolor{RefColor}{Strict token-wise streaming}}
\For{\rm{token} $t\in\mathcal{T}_b$}{
$(\hat z(t),\ell_t,\mathrm{KV})\leftarrow
f_\theta(x_t;\mathrm{KV},P,\alpha^{(b)})$\;
$P_e\leftarrow\hat z_{s_e}(t)-\hat z_{d_e}(t)\;(\forall e\in\mathcal{C})$\;
}
}
}
}
}
\end{algorithm}
Algorithm~\ref{alg:reflux} separates the three execution paths used in the
paper. In the calls above, \(P\) or \(p(t)\) supplies the payloads in
Eq.~\ref{eq:delta_write}, and \(\alpha^{(b)}\) supplies their block-wise write
strengths. Synchronous evaluation records a no-feedback reference traversal
and applies the resulting increments in a second traversal over the same block.
Strict streaming applies the recurrence throughout the scored sequence. For
generation, parallel-prefill streaming keeps the standard prompt prefill path
and activates the same payload cache only during autoregressive decoding.

\paragraph{KV-cache semantics.}
For synchronous language-modeling evaluation, each complete window is
processed twice with \texttt{use\_cache=False}: the first traversal records
the no-feedback reference states and the second applies their increments.
Autoregressive synchronous decoding instead maintains two independent KV
caches. At each step, the no-feedback traversal reads and advances only the reference cache;
the feedback traversal reads and advances only the feedback cache, and its
logits are used for prediction. Each trajectory therefore appends the current
token exactly once, and the reference states remain unaffected by earlier
feedback writes. Streaming decoding uses only the feedback cache, since its
payload at step \(t\) is taken from the feedback-augmented traversal at
\(t-1\).

\paragraph{Reference states and parallel prefill.}
In strict full-sequence streaming, the increment produced at position
$t{-}1$ is injected at position $t$, after which the increment at $t$ is
computed from the updated stream. This keeps the backbone computation at
$1.00\times$, but requires sequential processing and therefore does not
exploit window-level parallelism. Parallel-prefill streaming performs one standard batched prompt
traversal, retaining its KV cache and setting
\(P_e(T)=z_{s_e}^{(0)}(T)-z_{d_e}^{(0)}(T)\) at the final prompt position
\(T\). The first generated token consumes \(P_e(T)\); every subsequent
payload is then computed from the feedback-augmented decode state exactly as
in Eq.~\ref{eq:stream_payload}. This schedule leaves the prompt cache
unmodified and is evaluated separately from strict streaming in
Table~\ref{tab:prefill_decode_e2e}.

\paragraph{Multi-edge writes and payload cache.}
Active edges are applied at their target layers in increasing depth order;
the additive basis $h_d(t)$ of Eq.~\ref{eq:delta_write} is the current
stream at that depth, which already includes writes from edges targeting
shallower depths. The payload cache $P_e$ maintains slots for \emph{all}
candidate edges ($E \cdot D \cdot 2$ bytes;
App.~\ref{app:efficiency_protocol}) regardless of the current support, so an
edge newly activated at block $b{+}1$ always has a previous-position payload
available.

\paragraph{Causality.}
Writes at position $t$ depend only on states at positions $\le t$; future
tokens are neither read nor written, so a perturbation of a future token
leaves the logits of earlier positions unchanged by construction.

\section{Corpora and Datasets}
\label{app:datasets}

\subsection{Language-modeling corpora}

We select ten corpora to cover a broad range of registers over which
inference-time feedback may operate, from open-domain web text to highly
structured and formulaic documents.

\textbf{C4} \citep{raffel2020exploring} is a heuristically cleaned crawl of
web pages and represents the open-domain text that dominates pretraining.
\textbf{LAMBADA} \citep{paperno2016lambada} consists of novel passages whose
final word can only be reliably predicted from the full discourse context,
providing a probe of long-range coherence beyond local lexical statistics.
\textbf{arXiv} and \textbf{PubMed}, two components of The Pile
\citep{gao2020pile}, contribute scientific and biomedical text with dense
terminology, mathematical notation, and citation structure.
\textbf{PG19} \citep{rae2019compressive} contains long-form fiction published
before 1919, while \textbf{BookSum} \citep{kryscinski2022booksum} provides
book-length narrative text together with chapter- and book-level summaries;
together, they probe coherence over substantially longer spans.
\textbf{Newsroom} \citep{grusky2018newsroom} contains 1.3 million news
articles paired with summaries from 38 major outlets.
Finally, \textbf{BigPatent} \citep{sharma2019bigpatent}, comprising more than
one million granted US patents, \textbf{BillSum} \citep{kornilova2019billsum},
which covers US legislative bills, and \textbf{GovReport}
\citep{huang2021efficient}, which contains US government oversight and
research reports, provide long and highly structured documents with
sectional conventions that differ markedly from web text.
 
\subsection{Reasoning benchmarks}

The eight benchmarks cover factual knowledge, commonsense inference,
multi-step reasoning, and multi-hop composition.

\textbf{MMLU} \citep{hendrycks2020measuring} spans 57 subjects in a
four-choice multiple-choice format.
\textbf{ARC-Easy} and \textbf{ARC-Challenge} \citep{clark2018think}
consist of grade-school science questions, with the Challenge split designed
to exclude items solvable by simple retrieval or word-co-occurrence
baselines.
\textbf{HellaSwag} \citep{zellers2019hellaswag} contains adversarially
filtered sentence-completion problems drawn from physical and instructional
contexts, while \textbf{Winogrande} \citep{sakaguchi2021winogrande}
extends Winograd-style pronoun resolution through contrastive twin pairs.
\textbf{GSM8K} \citep{cobbe2021training} consists of grade-school
mathematical word problems requiring multi-step reasoning.

The final two benchmarks, \textbf{HotpotQA} \citep{yang2018hotpotqa} and
\textbf{2WikiMultiHopQA} \citep{ho2020constructing}, directly probe the
composition behavior targeted by our method. Both require multi-hop
question answering over Wikipedia-style passages, where relevant evidence
is distributed across documents and intermediate conclusions must remain
available for subsequent reasoning steps. This setting provides a
mechanistically informative test of ReFlux, as preserving intermediate
information across depth is precisely the behavior that our feedback
mechanism is designed to support.

\section{Prompt Templates}
\label{app:prompt_templates}

All models are evaluated as pretrained base language models without
instruction tuning or chat templates. We use the prompt formats of the
corresponding benchmark implementations, following the
widely adopted \texttt{lm-evaluation-harness} conventions. The same
prompt construction and decoding protocol are used for the base and
ReFlux variants. For few-shot evaluation, demonstrations are placed
before the target example using the benchmark's standard formatting.

\newcommand{\promptbox}[2]{%
\begin{tcolorbox}[
enhanced, breakable,
colback=BlueGreen!4,
colframe=BlueGreen!60!black,
boxrule=0.6pt,
arc=2.5pt,
left=6pt, right=6pt, top=6pt, bottom=5pt,
fontupper=\small\ttfamily,
title={\small\bfseries\sffamily #1},
coltitle=white,
colbacktitle=BlueGreen!60!black,
attach boxed title to top left={yshift=-2.5mm, xshift=4mm},
boxed title style={
colback=BlueGreen!60!black,
colframe=BlueGreen!60!black,
boxrule=0pt,
arc=1.5pt,
left=5pt, right=5pt, top=1.5pt, bottom=1.5pt
}
]
#2
\end{tcolorbox}}

\paragraph{MMLU.}

\vspace{0.3em}
We use the subject-specific multiple-choice format adopted by the
benchmark implementation. Each demonstration contains the question,
four answer choices, and the ground-truth answer, while the target
example omits the answer.

\promptbox{MMLU}{
The following are multiple choice questions (with answers) about
\textit{{subject}}.

\textit{{question}}

A. \textit{{choice A}} \
B. \textit{{choice B}} \
C. \textit{{choice C}} \
D. \textit{{choice D}}

Answer: \textit{{answer}}

\medskip

\textit{{target question}}

A. \textit{{choice A}} \
B. \textit{{choice B}} \
C. \textit{{choice C}} \
D. \textit{{choice D}}

Answer:
}

We use 5-shot evaluation. Predictions are obtained by comparing the
conditional likelihoods of the candidate answer labels, and we report
accuracy.

\paragraph{ARC-Easy.}
Following the \texttt{lm-evaluation-harness} task definition, the
question and answer prefix form the input prompt, while the candidate
choices are scored separately by the multiple-choice evaluation
interface.

\promptbox{ARC-Easy}{
Question: \textit{{question}}\
Answer:
}

We use 0-shot evaluation and select the candidate answer with the
highest conditional likelihood. We report accuracy.

\paragraph{ARC-Challenge.}
ARC-Challenge uses the same prompt construction as ARC-Easy, with a
different number of few-shot demonstrations.

\promptbox{ARC-Challenge}{
Question: \textit{{question}}\
Answer:
}

We use 25-shot evaluation and select the candidate answer with the
highest conditional likelihood. We report accuracy.

\paragraph{HellaSwag.}
For HellaSwag, the processed context constitutes the prompt, while the
candidate endings are evaluated as alternative continuations.

\promptbox{HellaSwag}{
\textit{{context / query}}
}

For each example, the model assigns a conditional likelihood to each
candidate ending, and the candidate with the highest score is selected.
We use 10-shot evaluation and report accuracy.

\paragraph{WinoGrande.}
WinoGrande is evaluated as a two-choice sentence-completion task. The
prompt contains the sentence with its blank, while the two candidate
fillers are scored as alternative continuations.

\promptbox{WinoGrande}{
\textit{{sentence with blank}}
}

We use 5-shot evaluation and select the candidate with the highest
conditional likelihood, and report accuracy.

\paragraph{GSM8K.}
We follow the 5-shot generation-based evaluation protocol provided by
\texttt{lm-evaluation-harness}. Each demonstration contains a question
and its reference solution, while the target example ends with an empty
answer field.

\promptbox{GSM8K}{
Question: \textit{{demonstration question}}\
Answer: \textit{{demonstration solution and final answer}}

\medskip

Question: \textit{{target question}}\
Answer:
}

The model generates a solution autoregressively with greedy decoding
(\texttt{temperature}=0), and the final numerical answer is extracted
from the generated response for exact-match evaluation. We use 5-shot
evaluation.

\paragraph{HotpotQA.}
For HotpotQA, we use a context-grounded multi-hop question-answering
format with the benchmark-provided context. The context is presented
before the question, and the model is asked to produce the final answer
directly.

\promptbox{HotpotQA}{
Answer the question based on the given paragraphs. 

The following are given paragraphs.

\textit{{context}}

Answer the question based on the given paragraphs. 

Question: \textit{{question}}\
Answer:
}

We use 0-shot evaluation without task-specific fine-tuning or external
retrieval, and report exact match (EM).

\paragraph{2WikiMultiHopQA.}

We use the same context-grounded question-answering format for
2WikiMultiHopQA.

\promptbox{2WikiMultiHopQA}{
Answer the question based on the given paragraphs. 

The following are given paragraphs.

\textit{{context}}

Answer the question based on the given paragraphs. 

Question: \textit{{question}}\
Answer:
}

We use 0-shot evaluation without task-specific fine-tuning or external
retrieval, and report exact match (EM).

\end{document}